\documentclass[acmsmall]{acmart}
\AtBeginDocument{%
  \providecommand\BibTeX{{%
    \normalfont B\kern-0.5em{\scshape i\kern-0.25em b}\kern-0.8em\TeX}}}

\setcopyright{acmcopyright}
\copyrightyear{2023}
\acmYear{2023}
\acmDOI{XXXXXXX.XXXXXXX}

\acmJournal{JACM}
\acmVolume{37}
\acmNumber{4}
\acmArticle{111}
\acmMonth{9}

\usepackage[utf8]{inputenc}
\usepackage{amsmath}
\usepackage{amsfonts}
\usepackage{color}
\usepackage{xcolor}
\usepackage{graphicx}
\def\ie{\textit{i.e.}}
\def\eg{\emph{e.g.}}

\usepackage{url}
\usepackage{multirow}
\usepackage{makecell}

\begin{document}

\title{Towards Data-centric Graph Machine Learning: Review and Outlook}

\author{Xin Zheng}
\authornote{Both authors contributed equally to this research.}
\email{xin.zheng@monash.edu}
\author{Yixin Liu}
\authornotemark[1]
\email{yixin.liu@monash.edu}
\affiliation{%
  \institution{Monash University}
  \city{Melbourne}
  \state{VIC}
  \country{Australia}
  \postcode{3800}
}

\author{Zhifeng Bao}
\affiliation{%
  \institution{RMIT University}
  \city{Melbourne}
  \state{VIC}
  \country{Australia}}
\email{zhifeng.bao@rmit.edu.au}

\author{Meng Fang}
\email{Meng.Fang@liverpool.ac.uk}
\affiliation{%
  \institution{University of Liverpool}
  \city{Liverpool}
  \country{UK}
}

\author{Xia Hu}
\email{xia.hu@rice.edu}
\affiliation{%
  \institution{Rice University}
  \city{Houston}
  \country{US}
}

\author{Alan Wee-Chung Liew}
\email{a.liew@griffith.edu.au}
\author{Shirui Pan}
\authornote{Correspoinding Author.}
\email{s.pan@griffith.edu.au}
\affiliation{%
 \institution{Griffith University}
 \city{Gold Coast}
 \state{Queensland}
 \country{Australia}}

\renewcommand{\shortauthors}{Zheng and Liu, et al.}

\begin{abstract}
  Data-centric AI, with its primary focus on the collection, management, and utilization of data to drive AI models and applications, has attracted increasing attention in recent years. In this article, we conduct an in-depth and comprehensive \textbf{review}, offering a forward-looking \textbf{outlook} on the current efforts in data-centric AI pertaining to graph data—the fundamental data structure for representing and capturing intricate dependencies among massive and diverse real-life entities. 
  We introduce a systematic framework, \textbf{Data-centric Graph Machine Learning (DC-GML)}, that encompasses all stages of the graph data lifecycle, including graph data collection, exploration, improvement, exploitation, and maintenance.
  A thorough taxonomy of each stage is presented to answer three critical graph-centric questions: 
   \textit{(1) how to enhance graph data availability and quality; (2) how to learn from graph data with limited-availability and low-quality; (3) how to build graph MLOps systems from the graph data-centric view.}
   Lastly, we pinpoint the future prospects of the DC-GML domain, providing insights to navigate its advancements and applications
  \footnote{Github Page: \href{https://github.com/Data-Centric-GraphML/awesome-papers}{https://github.com/Data-Centric-GraphML/awesome-papers}}.
  \end{abstract}



 \received{\today}

\maketitle

\newcommand{\bao}[1]{ {\color{blue}{Bao: #1}}}
\newcommand{\mf}[1]{ {\color{brown}{MF: #1}}}

\section{Introduction}
\begin{figure}[t]
    \centering
    \includegraphics[width=\textwidth]{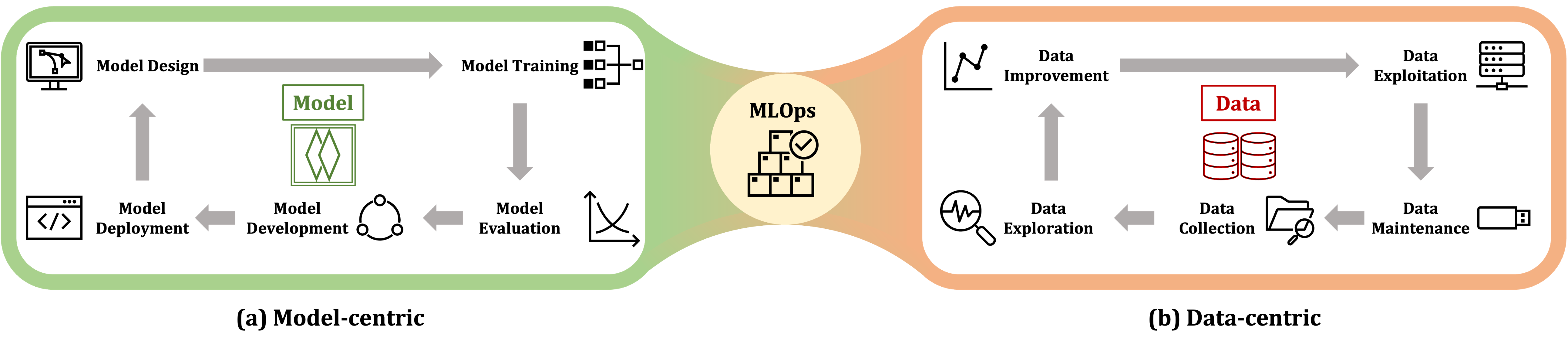}
    \caption{General comparison between (a) model-centric AI and (b) data-centric AI.}
    \label{fig:dcai_mcai}
\end{figure}

With the enormous growth of data, the advances and potentials of artificial intelligence (AI) have been explored and developed remarkably, creating ample opportunities for development in the research domain of machine learning (ML).
Over the years, researchers have dedicated their efforts to the development of model-centric AI, which has become a central focus in both domains of ML research and applications.
\textbf{Model-centric AI} aims to develop powerful models tailored to specific data for diverse learning tasks, with the assumption that the input data is well preprocessed/refined in a tidy format.
By placing a central focus on the advancement of models, well-designed models could achieve powerful learning performance.
However, in most if not all real-world applications, data are neither well-curated nor of high quality, especially when model development is a continuous process of learning from data. 
Recent statistics indicate that data scientists dedicate a significant portion, at least 80\%, of their time to data preparation~\cite{chai2023demystifying}. 
Unfortunately, the importance of preparing high-quality data in the entire AI system lifecycle has been largely overlooked in existing model-centric AI frameworks, as shown in Fig.~\ref{fig:dcai_mcai}~(a). 

Very recently, data-centric AI~\cite{budach2022effects,polyzotis2021can,polyzotis2019data} is attracting much attention from both academia and industry.
Different from the model-dominant objective in model-centric AI, \textbf{data-centric AI} aims to engineer data with great availability and quality for serving and promoting model-related ML tasks. There are shreds of evidence~\cite{Andrew-dc} showing that by enhancing data quality only, the machine learning models' performance on various tasks can be significantly improved, \eg, steel defect detection (from 76.2\% to 93.1\%) and solar panel inspection (from 75.7\% to 78.7\%)~\cite{Andrew-dc}.

A general comparison between model-centric AI and data-centric AI is illustrated in Fig.~\ref{fig:dcai_mcai}. 
As can be observed, model-centric AI focuses on the pipeline of model design~\cite{tedjopurnomo2020survey}, training~\cite{gao2022location}, evaluation~\cite{raschka2018model}, development~\cite{wu2020comprehensive}, and deployment~\cite{zaharia2018accelerating}.
In contrast, data-centric AI focuses on the comprehensive lifecycle of data, encompassing various stages of data engineering, involving data collection~\cite{whang2023data,chai2023demystifying}, exploration~\cite{ghorbani2019data,jia2019towards,luo2018deepeye}, improvement~\cite{li2021data,ding2022data}, exploitation~\cite{huang2023deep,liu2022graph}, and maintenance~\cite{renggli2019continuous,graphstorm}. 
In the overall process, models may not undergo further refinement until the corresponding input data is considered ready.
Hence, data-centric AI plays an important role in dealing with highly messy and noisy practical data in large quantities, and it serves well-processed data to guide model-centric AI for powerful model design.

Recently, two workshops~\cite{nipsw-21,stanfordw-21} in 2021 were initiated to discuss and explore data-centric AI, and two survey studies~\cite{zha2023dcaisurvey,zha2023data} were conducted to provide comprehensive data-centric AI blueprints.
All these developments indicate a noticeable shift in focus towards recognizing and prioritizing the significance of data, reflecting the growing interest in data-centric AI research. 
In contrast to existing surveys that generally discuss general data types in a broad perspective of data-centric AI, in this work, we aim to focus on debunking the mystery of data preparation for a typical data instantiation, \ie, graph-structured data.

Compared with other data types (\eg, images and texts), graph-structured data consists of discrete nodes connected by independent edges, reflecting its unique characteristics (\eg, sparsity and connectivity) in non-Euclidean space.
Such characteristics of graph-structured data make it particularly adept at representing complex structural relationships among massive diverse entities in the real world.
This remarkable representation capability of graph-structured data highlights the vital role of data-centric graph machine learning in advancing general data-centric AI.
In this work, we will extensively explore the field of data-centric graph machine learning (DC-GML) and provide comprehensive \textbf{review \& outlook} to navigate its advancements and applications.

\vspace{2mm}
\noindent{\textbf{Towards data-centric graph machine learning.}}
Taking graph-structured data as the central focus, \textbf{data-centric graph machine learning (DC-GML)} aims to process, analyze, and understand graph-structured data in its entire lifecycle, for enhancing the quality of graph data, uncovering significant insights from graph data, developing comprehensive representations of graph data, and working collaboratively with the graph model-centric development, under systematic graph machine learning operations (MLOps) pipeline.
In this work, 
we provide a comprehensive taxonomy that encompasses the progress of existing DC-GML works, and underscore the promising future prospects of DC-GML.
Concretely, we focus on three core questions towards DC-GML, 
and these core questions are indispensable yet fundamental components in a systematic graph MLOps workflow to realize DC-GML.
\begin{itemize}
\setlength{\parskip}{5pt}
    \item {Q1: How to enhance graph data availability and quality? 
    
    $\rightarrow$ (a) Graph data improvement.}
    \item {Q2: How to learn from graph data with limited-availability and low-quality? 
    
    $\rightarrow$ (b) Graph data exploitation.}
    \item {Q3: How to build graph MLOps systems from the graph data-centric view? 
    
    $\rightarrow$ (c) Graph data collection, (d) exploration, and (e) maintenance.}
\end{itemize}

\vspace{2mm}
\noindent{\textbf{Significance of data-centric graph machine learning (DC-GML).}} The critical role and necessity of working on DC-GML reflect in following aspects:

\vspace{1mm}
\noindent(1) \textit{Enhanced graph data understanding}: DC-GML explores underlying characteristics of complex and diverse graph data in real-world scenarios and addresses the inherent challenges posed by potentially messy and noisy graph data, through various processes and techniques, such as graph data visualization, graph data denoising, graph data valuation.

\vspace{1mm}
\noindent(2) \textit{Better graph learning model performance}: DC-GML improves the availability and quality of graph data, which accurately conveys useful and valuable information, as well as clear supervision signals that the model needs to learn. 
Such that graph learning models could effectively leverage high-quality graph data to capture complex patterns and make accurate predictions.

\vspace{1mm}
\noindent(3) \textit{Wider graph data application range}: With comprehensive graph data understanding and effective graph learning model development, DC-GML is able to model various open-world graph data for encouraging more application scenarios and insights, ranging from daily-life personal social networks to scientific chemical molecular study.

\vspace{1mm}
\noindent(4) \textit{Standardized graph machine learning workflow}: DC-GML enables a systematic graph MLOps framework by encompassing the entire graph data lifecycle. It offers consistent graph data collection, thorough exploration, continuous improvement, effective exploitation, and efficient maintenance, thereby benefiting significant cooperation between graph data-centric and model-centric development.

In summary, the contributions of this work can be listed as follows:
\begin{itemize}
\setlength{\parskip}{5pt}
    \item To the best of our knowledge, this is the first survey work in the field of data-centric AI specifically focused on graph-structured data.
    We provide a thorough landscape and promising prospects of data-centric graph machine learning (DC-GML) with an in-depth review of graph-centric research progress, which is expected to facilitate the understanding and practice of DC-GML and encourage future research outlooks.
    \item We provide a systematic taxonomy of DC-GML that encompasses all stages of the graph data lifecycle. This taxonomy includes graph data collection, exploration, improvement, exploitation, and maintenance, providing a comprehensive framework for DC-GML. Specifically, we aim to primarily answer two pivotal questions regarding the availability and quality of graph data in DC-GML, \ie,  how to enhance the graph data and how to effectively learn from it.
        
    \item Through the lens of graph machine learning model operations workflow (graph MLOps), we delve into various facets of graph data-centric research for answering the core question on how to build a graph MLOps from the graph data-centric view. Furthermore, we outline the open challenges and promising future directions in DC-GML, shedding light on areas that require further exploration and advancement.
\end{itemize}
The remaining parts of our survey are organized as follows.
Sec.~\ref{sec:background} describes necessary preliminaries, basic GNN backgrounds, and various downstream tasks of graph machine learning.
Then, Sec.~\ref{sec:roadmap} presents the proposed framework and taxonomy of DC-GML, providing a systematic review of existing progress.
Next, Sec.~\ref{sec:tasks}, Sec.~\ref{sec:paradigm}, and Sec.~\ref{sec:systems} comprehensively answer three core questions within five critical aspects of DC-GML.
Finally, Sec.~\ref{sec:futures} discusses potential challenges and promising future directions in DC-GML.

\section{Backgrounds}~\label{sec:background}
In this section, we provide backgrounds of data-centric graph machine learning. We describe the notations used in this paper, the core techniques of graph neural networks, and typical graph machine-learning tasks.

\subsection{Notations}

Throughout this paper, we use bold uppercase characters (\eg, $\mathbf{X}$) to denote matrices, bold lowercase characters (\eg, $\mathbf{x}$) to denote vectors, and calligraphic characters (\eg, $\mathcal{V}$) to denote sets. We use PyTorch-style indexing conventions for metrics and vectors. For instance, $\mathbf{X}[i,j]$ denotes the entry of matrix $\mathbf{X}$ at the $i$-th row and $j$-th column; $\mathbf{X}[i,:]$ and $\mathbf{X}[:,j]$ denotes the $i$-th row and $j$-th column of $\mathbf{X}$, respectively; and $\mathbf{x}[i]$ denotes the $i$-th entry of vector $\mathbf{x}$.

Let $G = (\mathcal{V}, \mathcal{E})$ denote a graph whose node set is $\mathcal{V}$ and edge set is $\mathcal{E}$. Let $v_i \in \mathcal{V}$ denote the $i$-th node and $e_{ij}=(v_i, v_j) \in \mathcal{E}$ denote the edge connecting $v_i$ and $v_j$. The number of nodes and the number of edges can be represented by $|\mathcal{V}|=n$ and $|\mathcal{E}|=m$, respectively. 
We can also represent the graph structure by an adjacency matrix $\mathbf{A} \in \{0,1\}^{n \times n}$, where $A[i,j]=1$ if $e_{ij} \in \mathcal{E}$ and $\mathbf{A}[i,j]=0$ if $e_{ij} \notin \mathcal{E}$. For a weighted graph where each edge $e_{ij}$ has an edge weight $w_{ij}$, the corresponding entry of its adjacency matrix $\mathbf{A} \in \mathbb{R}^{n \times n}$ is set to $\mathbf{A}[i,j]=w_{ij}$ for each edge. 
A graph may have node features that can be represented by a feature matrix $\mathbf{X} \in \mathbb{R}^{n \times d}$, where the $i$-th row $\mathbf{X}[i,:]$ denotes the feature vector of node $v_i$ and $d$ is the feature dimension. 
A graph may have node-level labels which can be represented as a label vector $\mathbf{y}$, where the $i$-th entry $\mathbf{y}[i]$ is the label of node $v_i$. 
Meanwhile, the graph-level label for a graph $G$ can be represented by $y_{G}$. 

\subsection{Graph Neural Networks}

Graph neural networks (GNNs) are a type of neural network architecture that operates on graph-structured data~\cite{kipf2017semi, wu2020comprehensive, jin2023survey}. Due to their strong expressive power, GNNs have achieved promising results on a wide range of graph learning tasks and become the de facto technique for graph machine learning~\cite{wu2020comprehensive, wu2019simplifying}. The mainstream GNNs follow the message passing paradigm that includes two core operations: {\em propagation} which aggregates the information from the neighboring nodes and {\em transformation} which models the node representations with non-linear mappings. Formally, the operation of a GNN layer can be represented by:

\begin{equation}
\begin{aligned}
\mathbf{m}_i^{(t)} & =  \operatorname{Propagate}\left(\mathbf{h}_i^{(t-1)},\left\{\mathbf{h}_j^{(t-1)} \mid v_j \in \mathcal{N}_{v_i}\right\}\right), \\
\mathbf{h}_i^{(t)} & =\operatorname{Transform}\left(\mathbf{h}_i^{(t-1)}, \mathbf{m}_i^{(t)}\right),
\end{aligned}
\end{equation}

\noindent where $\mathcal{N}_{v_i} = \{v_j |A[i,j] = 1\}$ is the neighbor set of node $v_i$, $\mathbf{h}_i^{(t)}$ is the representation vector of node $v_i$ at the $t$-th layer, $\mathbf{m}_i^{(t)}$ is the message vector of node $v_i$ at the $t$-th layer, $\operatorname{Propagate}(\cdot)$ and $\operatorname{Transform}(\cdot)$ are the propagation and transformation function, respectively.

To acquire a graph-level representation, a graph pooling layer is required to summarize compact information from the node-level representations of the graph. The operation of a graph pooling layer can be written by:

\begin{equation}
\mathbf{h}_G =\operatorname{Pool}\left(\left\{\mathbf{h}_i^{(T)} \mid v_i \in \mathcal{V}\right\}\right),
\end{equation}

\noindent where $\mathbf{h}_G$ is the representation vector of graph $G$, $\mathbf{h}_i^{(T)}$ is the node representation of $v_i$ at the final ($T$-th) layer, $\mathcal{V}$ is the node set of $G$, and $\operatorname{Pool}(\cdot)$ is the pooling function.

\subsection{Graph Machine Learning Tasks}

According to the scale of prediction targets, we divide the downstream tasks of graph machine learning into three categories, \ie, node-level tasks, edge-level tasks, and graph-level tasks. 

\vspace{1mm}\noindent\textbf{Node-level Tasks}.~ 
Node-level graph machine learning tasks aim to predict the specific property of each node in one or multiple graphs. Mainstream node-level tasks include node classification~\cite{kipf2017semi,velivckovic2018graph}, node regression~\cite{rong2020self}, and node clustering~\cite{pan2018adversarially}. As the most representative one, node classification aims to predict the discrete label for each node in a graph. To train a model for node classification, a cross-entropy loss is usually employed to minimize the difference between the predicted logits and the ground-truth labels of the training nodes. Differently, node regression aims to predict the continuous property of each node, while node clustering aims to divide nodes into different clusters without accessing their labels.  

\vspace{1mm}
\noindent\textbf{Edge-level Tasks}.~ 
Edge-level graph machine learning tasks focus on predicting the edge-level property of each edge in one or multiple graphs. Edge classification~\cite{aggarwal2016edge} and link prediction~\cite{zhang2018link} are two typical edge-level tasks. Taking link prediction as an example, it aims to discriminate the existence of the connection between two nodes. To this end, we usually regard link prediction as a binary classification problem, and a binary cross-entropy loss is commonly used to optimize link prediction models. Apart from link prediction, edge classification focuses on predicting the specific label of each edge rather than its existence.

\vspace{1mm}\noindent\textbf{Graph-level Tasks}.~ 
Unlike the above two tasks, graph-level tasks aim to model and make predictions at the level of the entire graph rather than individual nodes or edges. Graph-level tasks are often concerned with tasks such as graph classification~\cite{xu2018powerful}, graph regression~\cite{rong2020self}, graph generation~\cite{liao2019efficient}, and graph matching~\cite{bai2019simgnn}. Among them, graph classification is a fundamental graph-level task that involves assigning a label or a class to an entire graph. To optimize graph classification models, the cross-entropy between graph-level predictions and labels should be minimized. For other graph-level tasks, graph regression aims to predict the continuous property of different graphs, graph generation aims to generate new graph samples, and graph matching aims to measure the matching degree of a pair of graphs.

Overall, node-level, edge-level, and graph-level tasks form the core learning tasks of graph machine learning, enabling a comprehensive analysis and understanding of complex graph data. These tasks provide valuable insights and facilitate applications across diverse domains, including social network analysis, bioinformatics, recommendation systems, and many others.
\begin{figure}[t]
    \centering\includegraphics[width=0.85\textwidth]{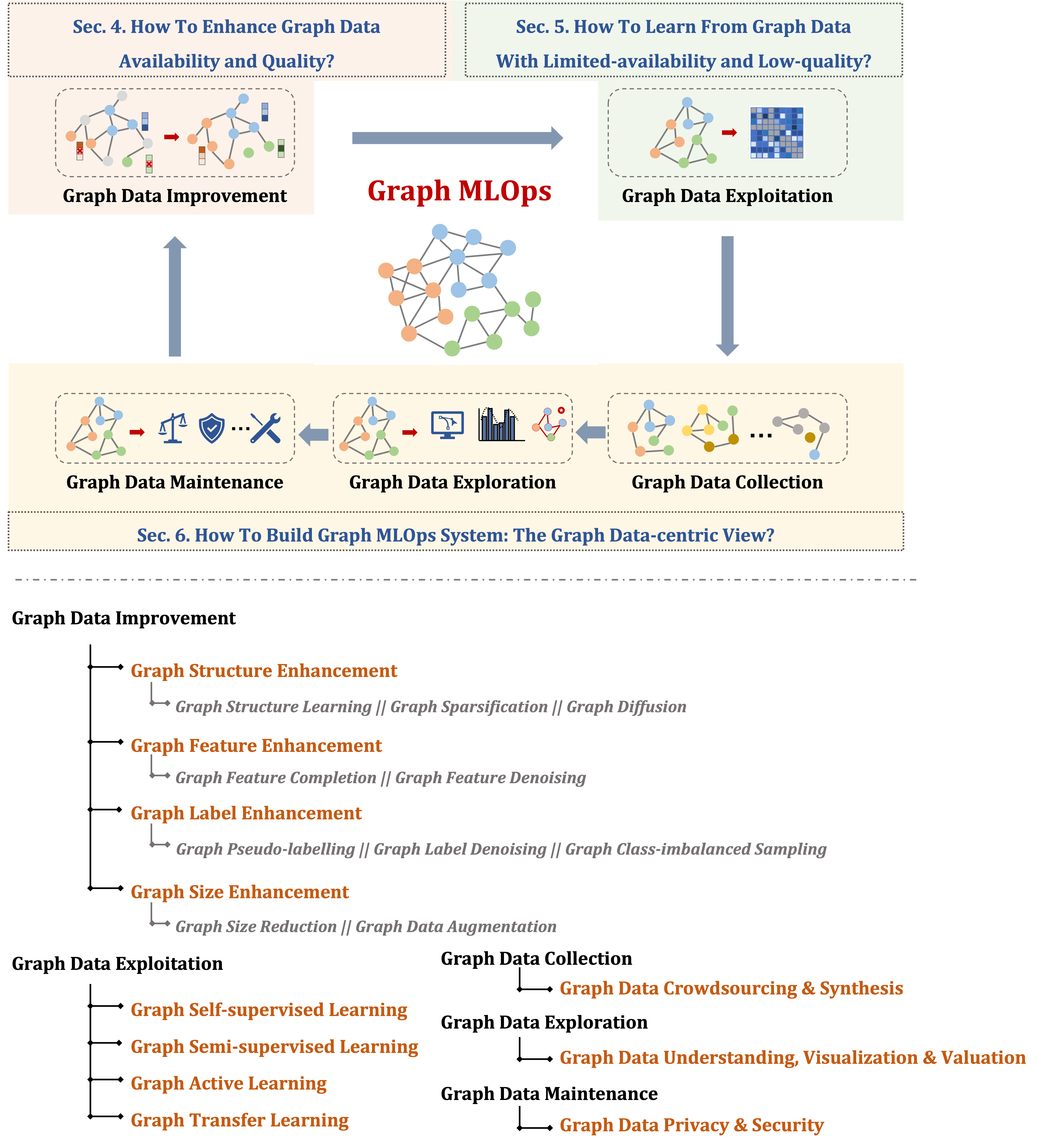}
    \caption{The framework and taxonomy of data-centric graph machine learning (DC-GML).}
    \label{fig:framework}
\end{figure}
\section{Framework \& Taxonomy}\label{sec:roadmap}
The proposed framework of data-centric graph machine learning (DC-GML) and the taxonomy of existing progress made in data-centric graph machine learning is illustrated in Fig.~\ref{fig:framework}. 

Concretely, we pay significant attention to two practical aspects concerning graph-structured data, namely \textit{availability} and \textit{quality}. 
Availability guarantees that there is sufficient graph data to be utilized for model development without scarcity, and quality ensures that there are no errors or noises in the graph data. 
Taking into account these factors, the three core questions that this article pays primary attention to are:
\begin{itemize}
\setlength{\parskip}{5pt}
\item {\textbf{Q1: How to enhance graph data availability and quality?}}
The answers to this question correspond to \textbf{graph data improvement} strategies in Sec.~\ref{sec:tasks}, which synthesize or modify graph data itself to improve availability and quality by comprehensively fixing potential issues of graph data. 
As shown in Fig.~\ref{fig:enhance}, the typical strategies contain: \textit{graph structure enhancement}, \textit{graph feature enhancement}, \textit{graph label enhancement}, and \textit{graph size enhancement}, by taking into account the characteristics of the graph data, covering graph structure, node/edge attribute features, and node/graph annotated labels, and graph data quantity.
\item {\textbf{Q2: How to learn from graph data with limited-availability and low-quality?}}
The answers to this question correspond to \textbf{graph data exploitation} strategies in Sec.~\ref{sec:paradigm}, which delve into scarce or low-quality graph data by fully extracting, exploiting, and integrating valuable information into graph machine learning models, when the graph data remains insufficient for effective model development even after experiencing graph data improvement. 
As shown in Fig.~\ref{fig:paradigm}, the typical strategies contain: \textit{graph self-supervised learning}, \textit{graph semi-supervised learning}, \textit{graph active learning}, and \textit{graph transfer learning}.
\begin{figure}[t]
    \centering
    \includegraphics[width=0.8\textwidth]{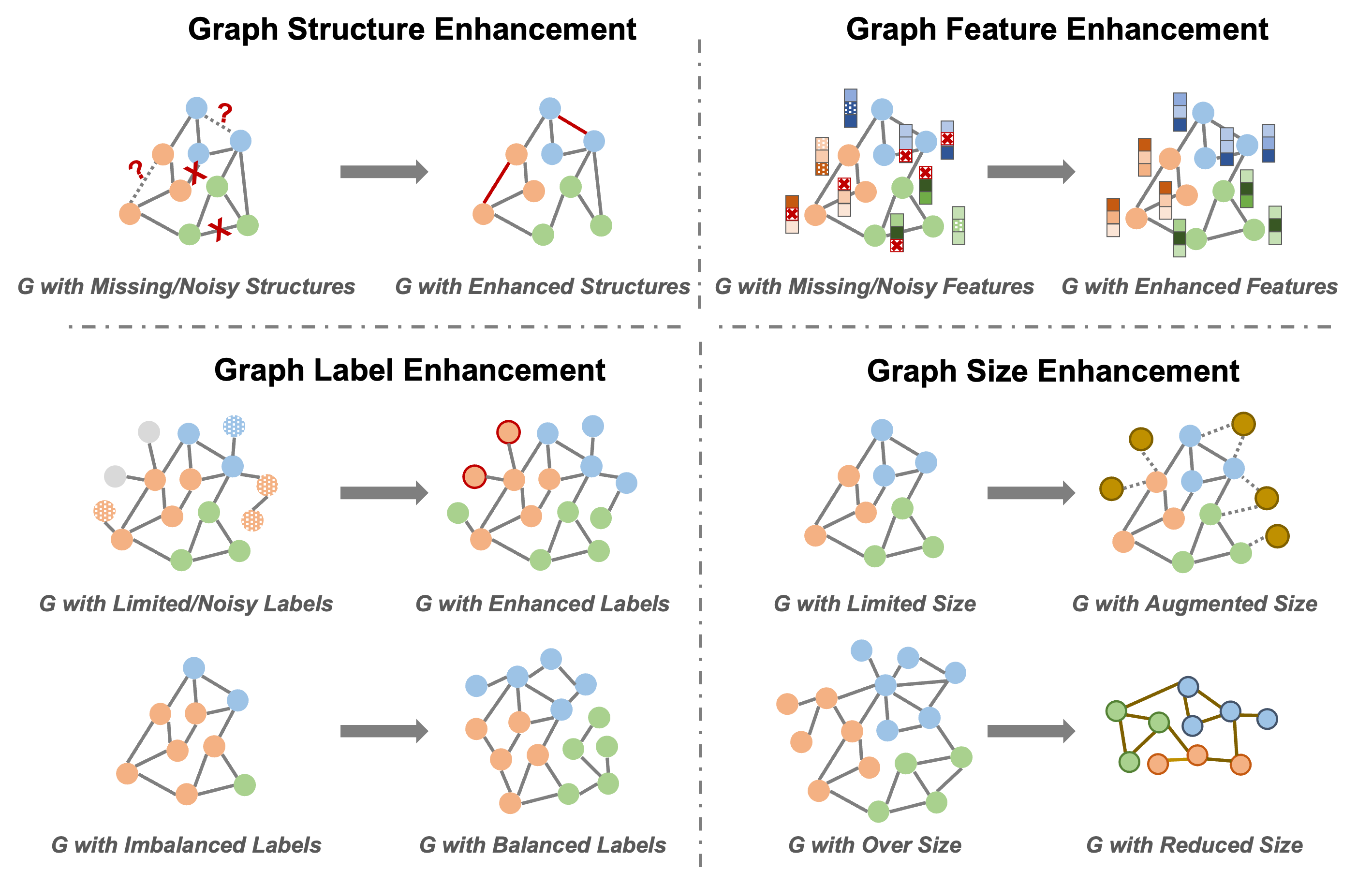}
    \caption{Illustration of graph data improvement strategies for effectively improving the availability and quality of graph data.}
    \label{fig:enhance}
\end{figure}

\begin{figure}[t]
    \centering
    \includegraphics[width=0.8\textwidth]{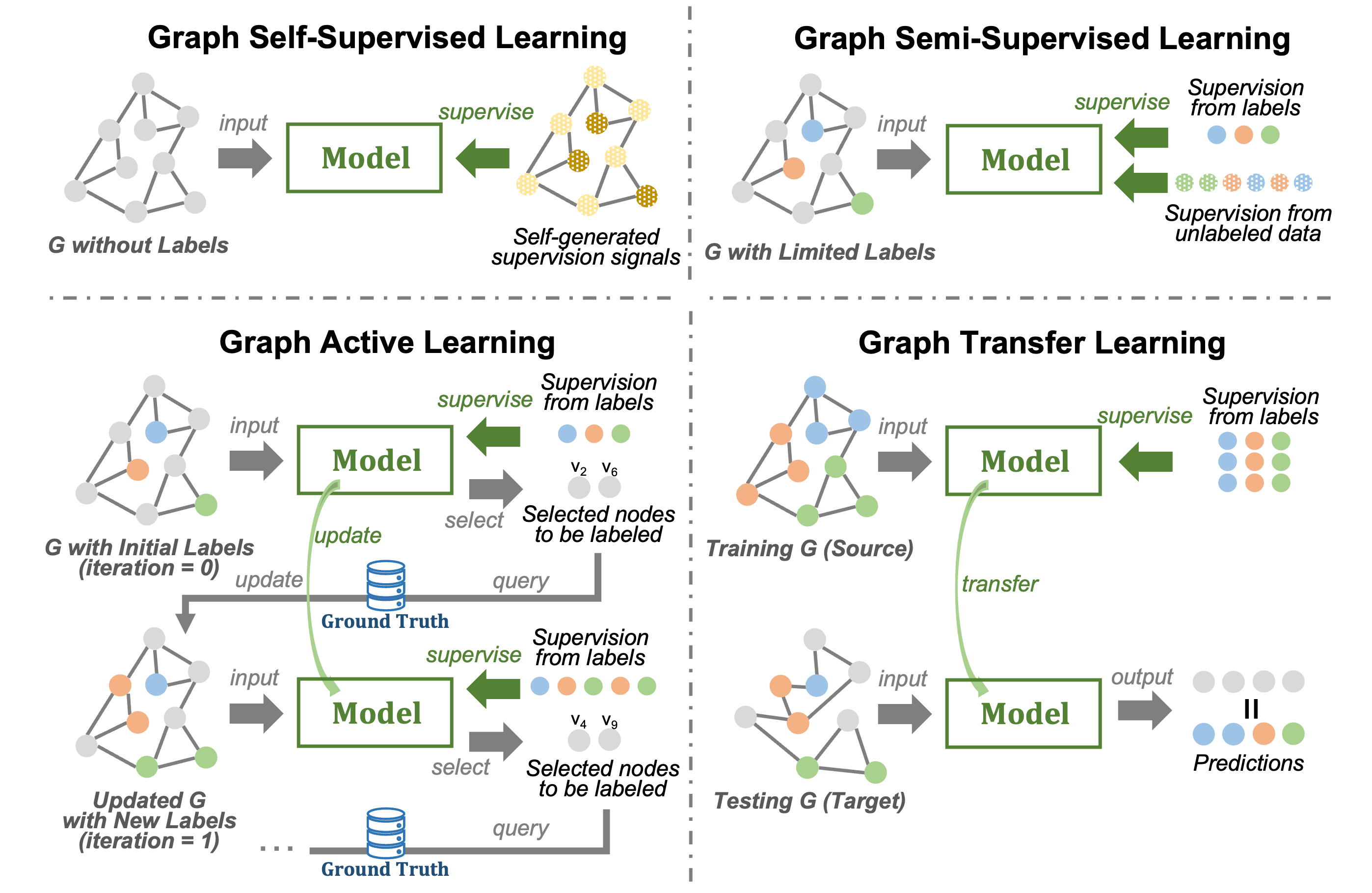}\caption{Illustration of graph data exploitation strategies for effectively learning from limited-availability and low-quality graph data.}
    \label{fig:paradigm}
\end{figure}
\item {\textbf{Q3: How to build graph MLOps systems from the graph data-centric view?}}
The answers to this question correspond to the strategies regarding the other three essential components in building a systematic graph MLOps workflow from the graph data-centric view, which cover:
\begin{itemize}
\item \textbf{Graph data collection}: use crowdsource
or synthesize graph data for providing sufficient supervision for graph machine learning model training;
\item  \textbf{Graph data exploration}: understand, analyze, and manage numerous and complex graph-structured data explicitly and comprehensively for graph machine learning model design;
\item  \textbf{Graph data maintenance}: maintain, update, and integrate in-service graph data under security and privacy management for graph machine learning model deployment.
\item  \textbf{Graph MLOps}: compose all the above critical components in graph data-centric lifecycle, together with model-centric graph machine learning for building the systematic graph MLOps workflow.
\end{itemize}
\end{itemize} 

\noindent\textbf{Connections among Q1, Q2, and Q3.} 
The answers to Q1 directly address and improve the availability and quality of graph-structured data.
However, even after undergoing significant enhancements, the graph-structured data may still fall short of being sufficient for the development of graph machine-learning models.
In this case, the answers to Q2 fully leverage valuable information with various learning strategies, under different scenarios with the limited availability and low quality of graph data, making it a bridge connecting current research on graph machine learning models towards the research centered on graph data engineering.
Taking a broader perspective, the answers to Q3 integrate all graph data-centric components that cover the entire graph data lifecycle and build a comprehensive graph MLOps system.
Therefore, these three core questions work collaboratively to explore the fundamental aspects of graph-centric data and drive progress towards the practical implementation of data-centric graph machine learning (DC-GML) within the proposed comprehensive framework.

\section{How To Enhance Graph Data Availability and Quality}\label{sec:tasks}
Enhancing graph data availability and quality is a fundamental aspect of data-centric graph machine learning, with its primary objective of graph data improvement.
Concretely, the core strategies aim to synthesize or modify graph data itself to improve availability and quality by comprehensively fixing potential issues of graph data. 
We consider three essential components of graph data, \ie, graph structure, node/edge attribute features, and node/graph annotated labels. 
For each component, we take into account two scenarios: limited availability with scarce or incomplete graph data and low quality with messy or noisy graph data.
Additionally, we consider the holistic quantity characteristic of graph data in two scenarios: oversized large-scale graphs with redundant information, and small-scale graphs with limited data sources and inadequate information.
More details about each strategy for graph data improvement are shown as follows.
\subsection{Graph Structure Enhancement}~\label{subsec:structure}
\begin{figure}[t]
    \centering
    \includegraphics[width=0.9\textwidth]{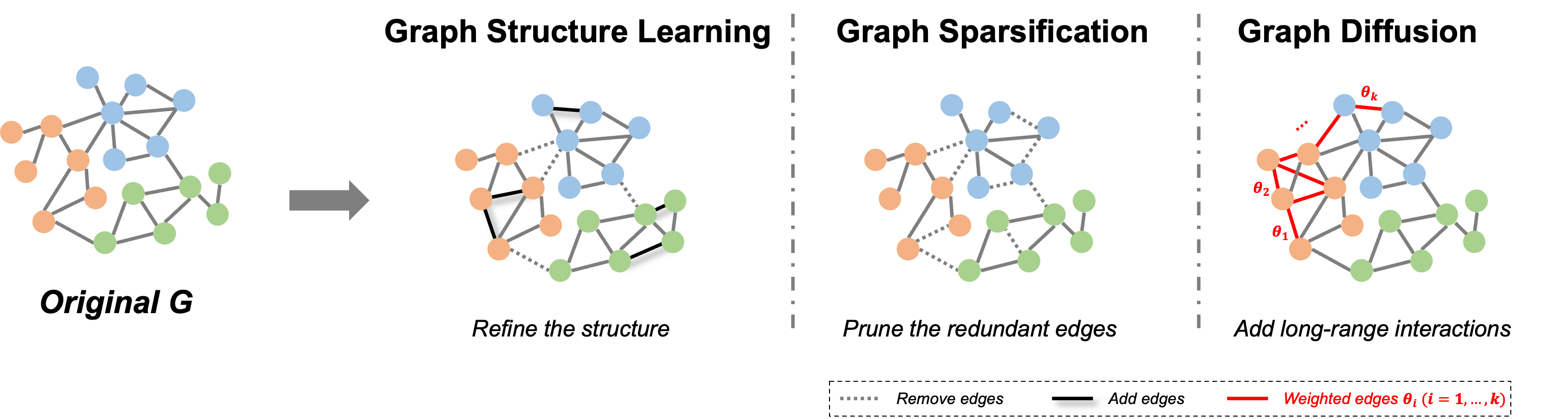}\caption{Illustration of graph structure enhancement methods.}\label{fig:structure_tasks}
\end{figure}

Different from image or text data where data points are embedded into regular grids, graph-structured data use irregular topology structures to represent the relationship among data points. 
Aiming to improve low-quality graph structures, various structure enhancement methods are formulated and studied by the community. 
In this subsection, we formulate and review three typical structure-based tasks: 
(1) \textit{graph structure learning} that targets to add, remove, and reweight the edges on noisy or incomplete structures; (2) \textit{graph sparsification} that aims to prune the redundant edges and generate sparse structures from over-dense structures; 
and (3) \textit{graph diffusion} that aims to establish links between disconnected nodes using pre-computed weights, enabling the explicit capture of global structural knowledge beyond the original local neighbor connections.
A brief illustration of three types of graph structure enhancement methods is demonstrated in Fig.~\ref{fig:structure_tasks}. The summary of typical methods in graph structure enhancement is presented in Table~\ref{tab:structure}.

\begin{table}[t]
\caption{Summary of graph data-centric structure enhancement methods. For the `Goals' column, `Generate' means to synthesize new graph structures, and `Refine' means to fix the noisy/incomplete/sparse/over-dense issues 
in the original graph structures. Brackets in `Goals' denote the graph properties preserved in the process of `Graph Sparsification'.}
\label{tab:structure}
\centering
\resizebox{\textwidth}{!}{
\begin{tabular}{p{4cm}p{4cm}ll}
\toprule
\begin{tabular}[c]{@{}l@{}}Graph Data-centric\\ Structure Enhancement\end{tabular}  & Methods                    & Techniques                                                    & Goals                       \\\midrule
\multirow{5}{*}{Graph Structure Learning} 
& LDS~\cite{franceschi2019learning}                        & Probabilistic modeling                                        & Generate/Refine                   \\
& GLCN~\cite{jiang2019semi}                       & Similarity-based metric learning                              & Refine                        \\
& Pro-GNN~\cite{jin2020graph}                     & Learnable adjacency modeling                                  & Generate/Refine                   \\
& IDGL~\cite{chen2020iterative}                       & Similarity-based metric learning                              & Generate/Refine                        \\
& GEN~\cite{wang2021graph}                        & Probabilistic modeling                                        & Refine                   \\\midrule
\multirow{5}{*}{Graph Sparsification}
& Satuluri et al.~\cite{satuluri2011local}   & Spectral sparsification                                       & Refine (Effective resistance) \\
& Spielman et al.~\cite{spielman2011spectral}  & Spectral sparsification                                       & Refine (Spectral)             \\
& Sadhanala et al.~\cite{sadhanala2016graph}& Spectral sparsification \& Sampling & Refine (Spectral)             \\
& Arora et al.~\cite{arora2019differentially}       & Random sampling                                               & Refine (Differential privacy) \\
& GraphSparsify~\cite{wickman2022generic}             & Learning-based sparsification                                 & Generate                   \\
\midrule
\multirow{5}{*}{Graph Diffusion}          
& PPNP~\cite{predict2019gasteiger}                       & Personalized PageRank (PPR)                                   & Refine                        \\
& APPNP~\cite{predict2019gasteiger}                      & Iterative PPR matrix approximation                            & Refine                        \\
& GDC~\cite{gasteiger2019diffusion}                        & PPR and heat kernel diffusion                                 & Refine                        \\
& GPR-GNN~\cite{adaptive2021chien}               & Adaptive PPR matrix learning                                  & Refine                        \\
& ADC~\cite{zhao2021adaptive}                        & Automatic propagation estimation                              & Refine                       \\
\bottomrule
\end{tabular}
}
\end{table}

\subsubsection{Graph Structure Learning}

In real-world data, the graph structures are usually extracted from complex interaction systems, which inevitably contain uncertain, redundant, wrong, and missing connections. Aiming to tackle these issues, graph structure learning (GSL) is a promising technique that learns to construct and refine a more reliable graph topology from the original one. 
Modern GSL approaches are usually based on GNNs, and follow a joint learning pipeline, \ie, the graph topology is jointly optimized along with the GNN model under the supervision of downstream tasks. To make the graph structure optimizable, GSL approaches usually model the graph adjacency matrix with learnable parameters in different manners. 

A straightforward strategy is \textit{full adjacency modeling}, where each element in the adjacency matrix is modeled by an independent learnable parameter. For instance, Pro-GNN~\cite{jin2020graph} constructs a learnable adjacency matrix based on the original one, and jointly optimizes the learnable adjacency matrix and GNN parameters during model training. Full adjacency modeling methods enjoy considerable flexibility in constructing topology, since the weights of every potential edge are independent. However, independent modeling also brings large ($\mathcal{O}(n^2)$) memory requirements, resulting in the difficulty of modeling large-scale graphs. Moreover, with full adjacency modeling, some ideal properties of graph structure (\eg, smoothness and sparsity) are hard to preserve. To keep these key properties, existing methods usually introduce extra loss terms to regularize the learned adjacency matrix, such as sparsity regularization, feature/label smoothness regularization, and low-rank regularization. 

In consideration of the discrete nature of graph topology, \textit{probabilistic modeling}, another type of GSL methods, uses parameterized probabilistic models to construct the node dependency in graph-structured data. For example, LDS~\cite{franceschi2019learning} uses Bernoulli probability model to simulate the edges, and GEN~\cite{wang2021graph} leverages the stochastic block model for graph construction. A significant advantage of probabilistic modeling approaches is that the discrete and sparse properties of learned graphs can be preserved. However, the sampling process makes the parameters of probabilistic models hard to be optimized. To address this problem, approximation, and reparameterization tricks are usually considered in probabilistic modeling approaches.

On the basis of homophily assumption, \textit{metric learning} GSL models node connections with node-wise similarity computed by metric learning functions. For instance, GLCN~\cite{jiang2019semi} considers the node-wise difference as the metric function, where the similarity is calculated from the low-dimensional projection of raw features. IDGL~\cite{chen2020iterative} generates topology from the cosine similarity of both raw features and learned embeddings, and then integrates the learned structure with the original one. To acquire sparse graph structures, metric learning-based methods apply sparsification operations to the dense similarity matrix, such as kNN sparsification and threshold-based sparsification. Compared to the aforementioned types of methods, metric learning GSL approaches usually need fewer parameters for adjacency modeling and are easier to train. Nevertheless, for large-scale graphs with millions of nodes, how to efficiently calculate the pair-wise similarity is also a challenging problem. 

\subsubsection{Graph Sparsification}
Graph sparsification aims to reduce the number of edges in over-dense graphs while preserving important structural properties~\cite{batson2013spectral}, so that the graph structures can be significantly refined and improved with high data quality and model computation.
Graph sparsification methods serve a wide range of real-world graph learning applications, such as Laplacian smoothing, spectral clustering, and graph partitioning.

Typically,~\citet{sadhanala2016graph} considered three types of graph sparsification methods -- spectral sparsification method, uniform sampling method, and k-neighbors sampling method -- to alleviate the computational complexity of Laplacian smoothing. 
The core idea is to preserve the important properties of the Laplacian matrix in the sparsification process.
Concretely, spectral sparsification method~\cite{spielman2011spectral} constructs a sparser graph by preserving the spectral properties of Laplacian eigenvalues and eigenvectors in the original graph, reflecting as the effective resistance~\cite{satuluri2011local}.
In contrast, uniform sampling method is more straightforward by randomly selecting edges for sparsifying a graph.
When graph structures are complex and imbalanced, such a uniform sampling method would fail to construct a sparser graph that is highly closest to the original graph.
Moreover, k-neighbors sampling method considers the distances between nodes and preserves the local connectivity of the original graph by only keeping edges with k nearest neighbors for graph sparsification. 
Overall, uniform sampling is faster, but it cannot guarantee the preservation of specific graph properties that are closest to those of the original graph.
When graph structures are not too complex, k-neighbors sampling could achieve comparable performance to spectral sparsification with remarkable robustness.
Despite that different graph properties are preserved in the process of sparsification, existing methods generally follow the above three patterns for graph sparsification on different graph learning tasks. 
For example,~\citet{arora2019differentially} targets the graph sparsification in the random sampling pattern while ensuring differential privacy.

Different from the above conventional spectrum-based or sampling-based methods, GraphSparsify~\cite{wickman2022generic} introduces the learning-based graph sparsification method, to adaptively learn a sparsification strategy that preserves the key properties of the original graph.
The main idea is to develop models to learn the importance of edges for the graph structure certain criteria~\cite{zheng2020robust}.
Specifically, GraphSparsify uses a deep reinforcement learning agent to select edges to retain in the sparser graph, and it uses a graph environment that simulates the effects of edge removal on the graph's properties, such as its Laplacian matrix, connectivity, and clustering coefficient. 
Learning-based graph sparsification method offers the advantage of greater adaptability, allowing for flexible adjustments to various sparsification tasks and criteria in graph learning.

\subsubsection{Graph Diffusion}
Graph data usually exhibits the sparsity property, meaning that each node is only connected with a small number of nodes. With such a property, original graph structures explicitly include the local structural knowledge, \ie, the information of the closest neighbors. However, the global structural knowledge, \ie, the long-range interaction between nodes and the global roles of nodes, is usually concealed beneath the original structures. Aiming to solve this problem, graph diffusion is a simple yet effective solution that explicitly captures global structural knowledge. 

The core idea of graph diffusion is to reform the graph adjacency matrix into a diffusion adjacency matrix $\mathbf{S}$ where not only adjacent nodes but also long-range nodes are connected. In a diffusion adjacency matrix, each entry (\ie, the edge weight) is a continuous value $\mathbf{S}[i,j] \in [0,1]$ that indicates the connection strength between nodes $v_i$ and $v_j$. Formally, a generalized graph diffusion can be written by:

\begin{equation}
\mathbf{S}=\sum_{k=0}^{\infty} \theta_k \mathbf{T}^k,
\end{equation}

\noindent where $\theta_k$ is the weighting coefficient and $\mathbf{T}$ is the generalized transition matrix derived from adjacency matrix $\mathbf{A}$. Here, $\theta_k$ should satisfy $\sum_{k=0}^{\infty} \theta_k=1, \theta_k \in[0,1]$ and the eigenvalues of $\mathbf{T}$ should be bound by $\lambda_i \in[0,1]$. Two popular implementations of graph diffusion are personalized PageRank (PPR) and heat kernel. PPR applies $\mathbf{T} = \mathbf{A}\mathbf{D}^{-1}$ and $\theta_k=\alpha(1-\alpha)^k$ where $\alpha \in(0,1)$ is the teleport probability. Heat kernel defines $\mathbf{T} = \mathbf{A}\mathbf{D}^{-1}$ and $\theta_k=e^{-t} \frac{t^k}{k !}$ where $t$ is the diffusion time. Theoretically, the closed-form solutions of PPR and heat kernel can be directly computed; however, for large-scale graph data, estimating the graph diffusion matrix with finite steps of iteration number $k$ is more practical due to the heavy computational cost of closed-form solutions. 

To enable long-range interaction in graphs, PPNP~\cite{predict2019gasteiger} uses the PPR diffusion matrix to execute the information propagation. To Apply PPNP on larger graphs, APPNP~\cite{predict2019gasteiger} is an advanced method that approximates the PPR matrix with its iterative solution. GDC~\cite{gasteiger2019diffusion} well formulates the graph diffusion process and replaces the conventional graph convolution in GNNs with PPR and heat kernel diffusion matrices. Beyond graph kernel with a set of fixed weighting coefficients, Chien et al.~\cite{adaptive2021chien} propose to adaptively learn the coefficients in a PPR-based diffusion framework. Similarly, ADC~\cite{zhao2021adaptive} uses an automatic estimation strategy to find the optimal propagation in graph diffusion-based GNNs.

\begin{table}[t]
\caption{Summary of graph data-centric feature enhancement methods. In the `Targets' column for feature completion, `Incomplete' means to complete the non-empty feature sets for every node, and `Missing' means to  complete the entirely missing feature sets for partial nodes. In the `Categories' column for feature denoising, methods can be categorized into `Graph signal processing (SP)-based' groups and `GNN-based' groups.}
\label{tab:feature}
\centering
\resizebox{\textwidth}{!}{
\begin{tabular}{p{4cm}p{4cm}p{7cm}l}
\toprule
\begin{tabular}[c]{@{}l@{}}Graph Data-centric\\ Feature Enhancement\end{tabular} & Methods                                 & Techniques                              & Targets/Categories             \\\midrule
\multirow{6}{*}{Graph Feature Completion}                                         
& GINN~\cite{spinelli2020missing}         & GNN-based autoencoder                   & Incomplete                    \\
& SAT~\cite{chen2020learning}             & Latent space shared autoencoders        & Missing                       \\
& GCN$_\text{MF}$~\cite{taguchi2021graph} & Gaussian mixture model                  & Incomplete                    \\
& HGNN-AC~\cite{jin2021heterogeneous}     & Pre-learned topological embedding        & Missing                       \\
& SAGA~\cite{tu2021siamese}               & Siamese network                         & Missing                       \\
& Amer~\cite{jin2022amer}                 & Neural network-based generator          & Missing                       \\
\midrule
\multirow{8}{*}{Graph Feature Denoising}                                          
& Shuman et.al ~\cite{shuman2013emerging} & Graph signal processing                 & Graph SP-based \\
& Chen et al.~\cite{chen2014signal}       & Graph total variation regularization    & Graph SP-based \\
& Wang et al.~\cite{wang2015trend}        & Graph trend filtering                   & Graph SP-based \\
& Waheed et al.~\cite{waheed2018graph}    & Graph polynomial filter                 & Graph SP-based \\
& Do et al.~\cite{do2020graph}            & Kron reduction-based graph autoencoder  & GNN-based                     \\
& GUN~\cite{chen2021graph}                & Graph unrolling networks                & GNN-based                     \\
& Rey et al.~\cite{rey2022untrained}      & Untrained graph-convolutional generator & GNN-based                     \\
& MAGNET~\cite{zhou2023robust}            & Sparsity-promoted graph autoencoder     & GNN-based         \\
\bottomrule           
\end{tabular}
}
\end{table}
\subsection{Graph Feature Enhancement}~\label{subsec:feature}

\begin{figure}[t]
    \centering
    \includegraphics[width=0.9\textwidth]{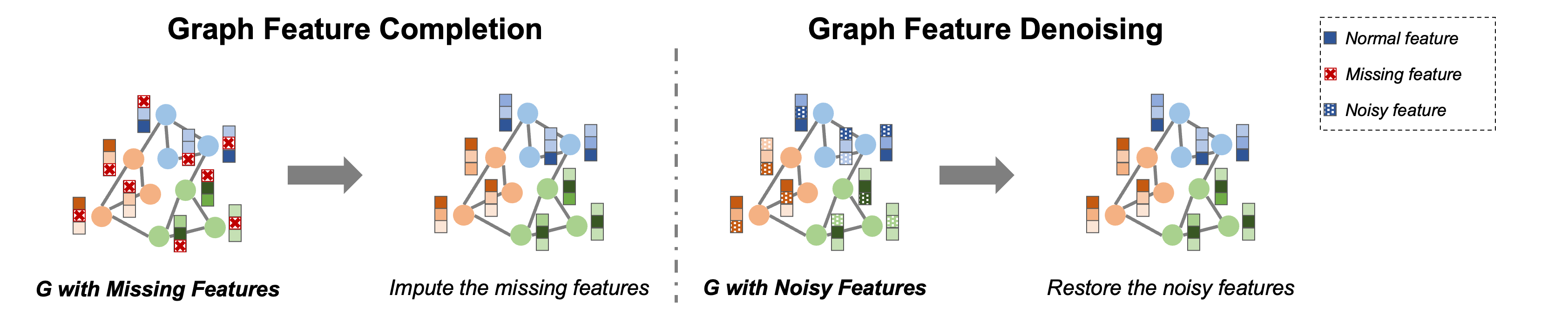}
    \caption{Illustration of graph feature enhancement methods.}
    \label{fig:feature_tasks}
\end{figure}

In graph-structured data, features are an essential element that represents the semantic information of each node and/or edge. For instance, in citation networks, node features are represented as bag-of-word vectors capturing the content of each paper, which offers valuable semantic knowledge for downstream tasks. However, these graph features in real-world datasets often suffer from low quality. For example, in a citation network, the extracted features may miss important words or include irrelevant ones, resulting in situations of missing features and noisy features. To enhance the quality of features in graph-structured data, two basic data-centric tasks have been widely investigated in recent years: (1) \textit{graph feature completion} that focuses on imputing the missing features and (2) \textit{graph feature denoising} that focuses on refining the noisy features. In this subsection, we summarize the existing works on these two tasks respectively. An illustration of two types of graph feature enhancement methods is shown in Fig.~\ref{fig:feature_tasks}, and the summary of typical methods of graph feature enhancement is presented in Table~\ref{tab:feature}. 

\subsubsection{Graph Feature Completion}

Graph feature completion aims to fill up the missing features in graph-structured data. Due to privacy protection or other data issues, the features of real-world graphs are sometimes incomplete or missing, which results in the low quality of graph data. In this case, GFC is a critical solution to improve the quality of features. According to the target of feature completion, existing GFC methods can be divided into two categories: \textit{incomplete feature completion} that complete the non-empty feature sets for every node and \textit{missing feature completion} that complete the entirely missing feature sets for partial nodes. 

For incomplete feature completion, an early work that applies GNNs for feature imputation is GINN~\cite{spinelli2020missing}. The core idea of GINN is to construct incomplete features with a GNN-based autoencoder network. To speed up the training process, GINN further employs an adversarial loss that learns to distinguish between imputed and original data. 
Aiming to process missing features and learning GNN with a uniform model, Taguchi et al.~\cite{taguchi2021graph} propose GCN$_\text{MF}$ which utilizes a Gaussian Mixture Model (GMM) to impute the incomplete data before the first hidden layer of GCN. The learning target of node classification provides supervision signals for model training.  

Among the methods for missing feature completion, SAT~\cite{chen2020learning} is a pioneering method. SAT uses two autoencoder models with a shared-latent space to model the structure and feature information respectively. The missing features can be imputed by a feature-based autoencoder. An adversarial distribution matching mechanism is leveraged to align the latent representations at two models. To complete attributes for heterogeneous graphs, HGNN-AC~\cite{jin2021heterogeneous} introduces a pre-learned topological embedding to guide the feature imputation from neighboring nodes. Amer~\cite{jin2022amer} utilizes a neural network-based generator for attribute completion, and a mutual information maximization strategy is used to optimize the model. 
SAGA~\cite{tu2021siamese} employs a siamese network to embed feature and structure information jointly, where a GNN-based decoder is used for data imputation. 

\subsubsection{Graph Feature Denoising}
In graph-structured data, features are usually collected from complex systems. Since real-world systems inevitably contain some uncertain or wrong information, the features are sometimes noisy and unreliable, which severely degrades the quality of graph data. To mitigate this issue, graph feature denoising aims to fix and refine noisy feature information to make the graph data more reliable. Early works utilize the technique of graph signal processing~\cite{shuman2013emerging}. This type of methods regards feature information as graph signals and then uses specialized graph filters to perform graph signal denoising. With the flourishing development of graph deep learning, another branch of methods conducts graph feature denoising with well-designed GNN models. 

In graph signal processing-based feature denoising methods, the node features are defined as graph signals~\cite{shuman2013emerging}. Then, the noisy features can be viewed as noisy graph signals composed of true signals and noise~\cite{chen2014signal}. To separate the noiseless graph signals from noise, a series of graph signal denoising models with different learning objectives and graph filters are proposed. For instance,~\citet{chen2014signal} propose an algorithm based on the regularization of graph total variation of noisy signals for graph signal denoising. In the proposed algorithm, both a closed-form solution derived from inverse graph filters and an iterative solution derived from standard graph filters are available for graph signal denoising implementation. Following the target of graph signal denoising, Waheed et al.~\cite{waheed2018graph} leverage a graph polynomial filter to approximate the eigenvalues of the inverse, which reduces the computational complexity of graph signal filtering. Apart from the above filters, graph trend filtering~\cite{wang2015trend} is also proven to be effective for graph signal denoising. To sum up, the key idea of graph signal processing-based methods is to design various graph filters to remove the noisy signals from observed graph signals, \ie, node-level features.

In view of the theoretical connections between graph signal processing and spectral GNNs~\cite{kipf2017semi,ma2021unified}, another line of work tries to leverage the more powerful GNNs for feature denoising. For instance, Do et al.~\cite{do2020graph} propose a graph autoencoder with Kron reduction-based pooling and unpooling to execute effective graph feature denoising on real-world traffic data. Equipped with an edge-weight-sharing graph convolution operation, Graph Unrolling Networks~\cite{chen2021graph} conduct an iterative denoising of graph signals along with the feed-forward process of each GNN layer.~\citet{rey2022untrained} propose untrained GNNs for graph signal denoising with a specially designed graph-convolutional generator network. MAGNET~\cite{zhou2023robust} uses a graph autoencoder to detect locally corrupted feature attributes in a graph, and then recovers a robust embedding for prediction tasks by pinpointing the positions of anomalous node attributes in an unbiased mask matrix and using a sparsity-promoting regularizer to recover robust estimations.
\begin{table}[t]
\caption{Summary of graph data-centric label enhancement methods.}
\label{tab:label}
\centering
\resizebox{\textwidth}{!}{
\begin{tabular}{p{5.5cm}p{3cm}ll}
\toprule
\begin{tabular}[c]{@{}l@{}}Graph Data-centric \\ Label Enhancement\end{tabular} & Methods                                                        & Techniques                                       & Categories                    \\\midrule
\multirow{6}{*}{Graph Pseudo-labeling}                                          
& Li et al.~\cite{li2018deeper}            & top-K high-confidence with self-training         & Selection-based              \\
& DSGCN~\cite{zhou2019dynamic}             & Threshold-based soft label confidence            & Selection-based              \\
& Sun et al.~\cite{sun2020multi}           & Feature clustering pseudo-labeling               & Generation-based             \\
& IFC-GCN~\cite{hu2021rectifying}          & Feature clustering pseudo-labeling               & Generation-based             \\
& RS-GNN~\cite{dai2022towards}             & Edge weight-based pseudo-labeling                & Generation-based             \\
& InfoGNN~\cite{li2023informative}         & Mutual information based pseudo-labeling         & Selection-based              \\
\midrule
\multirow{5}{*}{Graph Label Denoising}
& D-GNN~\cite{nt2019learning}              & Label noise estimation \& correction             & Metric-based                 \\
& IFC-GCN~\cite{hu2021rectifying}          & Expectation-Maximization rectification           & Surrogate loss-based         \\
& NRGNN~\cite{dai2021nrgnn}                & Correctly-labeled nodes relinking                & Metric based.                 \\
& CLNode~\cite{wei2023clnode}              & Global difficulty measurement                    & Metric-based                 \\
& RTGNN~\cite{qian2023robust}              & Self-reinforcement \& consistency regularization & Surrogate loss-based         \\
\midrule
\multirow{10}{*}{Graph Class-imbalanced Sampling}
& igBoost~\cite{pan2013graph}              & Ensemble learning with subgraph features         & Graph-level        \\
& GraphSMOTE~\cite{zhao2021graphsmote} & Node embedding based synthesis                   & Node-level (Minority)\\
& ImGAGN~\cite{qu2021imgagn}               & Minority generator                               & Node-level (Minority)         \\
& PC-GNN~\cite{liu2021pcgnn}               & Label-balanced neighborhood sampler              & Node-level (Majority)         \\
& GraphMixup~\cite{wang2021mixup}          & Node/Edge embedding mixup                        & Node-level (Minority)         \\
& GraphENS~\cite{park2021graphens}         & Node neighbor mixing                             & Node-level (Minority)         \\
& ReNode~\cite{chen2021topology}           & Topology-Imbalance Learning (TIL)                & Structure-level \\
& TopoImb~\cite{zhao2022topoimb}           & Topology extractor for sub-class imbalance       & Structure-level \\
& G2GNN~\cite{wang2022imbalanced}          & Neighboring graphs augmentation                  & Graph-level\\
& GraphSR~\cite{zhou2023graphsr}           & Unlabelled node based synthesis                  & Node-level (Minority)         \\
\bottomrule        
\end{tabular}
}
\end{table}
\begin{figure}[t]
    \centering
    \includegraphics[width=0.95\textwidth]{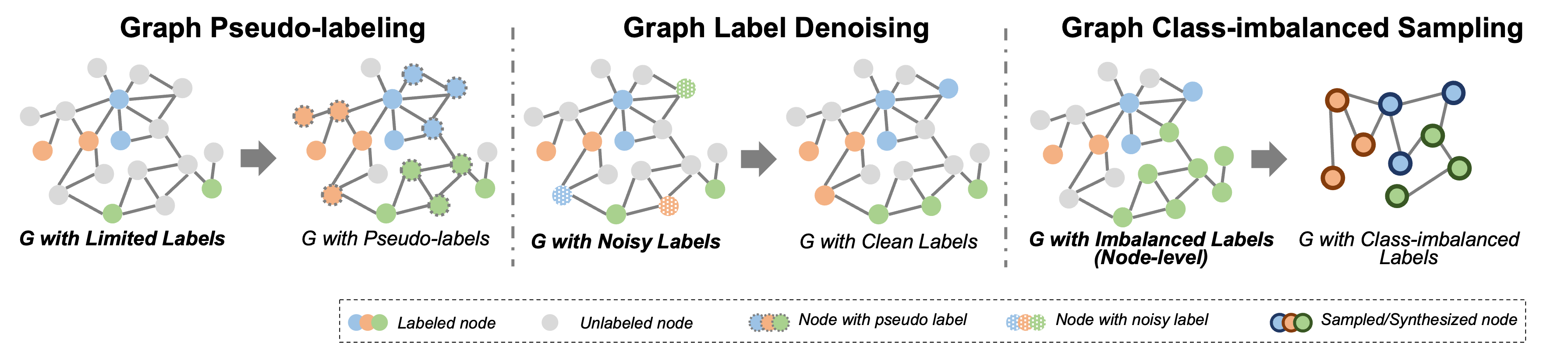}
    \caption{Illustration of graph label enhancement methods.}
    \label{fig:label_tasks}
\end{figure}
\subsection{Graph Label Enhancement}~\label{subsec:label}
As an essential component of graph-structured data under supervised learning, labels guide us in better analyzing and understanding graph data for benefiting the model development with accurate predictions.
On graph-structured data, labels usually mean the node class labels or the entire graph class labels. 
We consider three scenarios related to labels in data-centric graph machine learning: 
(1) \textit{scarce labels}, when massive graph data lacks node-level and graph-level class annotations as adequate information; 
(2) \textit{noisy labels}, when labels in graph data are not accurate enough to supervise the training of graph  learning models; 
(3) \textit{class-imbalanced labels}, when the number of examples for each class is imbalanced, the graph learning model performance might have severe bias.

The above three challenging graph data label-based issues could be addressed by three typical graph-centric tasks, \ie, 
(a) \textit{graph pseudo-labeling} for enriching the label information to alleviate the \textit{scarce label} issue; 
(b) \textit{graph label-denoising} for removing the redundant noisy label information to clean the \textit{noisy label} issue.
(c) \textit{graph class-imbalanced sampling} for downsampling majority and/or synthesizing minority class labels to tackle the \textit{class-imbalanced label} issue.
The sketch map of three types of graph label enhancement methods is given in Fig.~\ref{fig:label_tasks}, and a summary of methods in graph label enhancement is listed in Table~\ref{tab:label}.

\subsubsection{Graph Pseudo-labeling} As important supervision signals, labels of nodes and graphs are very important for developing expressive graph machine learning models with discriminative and representative node and graph representations. 
However, when collecting graph data in the real world, labels are usually limited, as obtaining more labeled nodes is time-consuming and expensive~\cite{dai2022towards}.
In this way, popular GNN architectures would become ineffective in propagating with the limited training labels, resulting in inferior performance.
In this section, we consider pseudo-labeling, a  complementary solution to address label scarcity on graph structure data~\cite{li2023informative}.
Concretely, graph pseudo-labeling aims to augment the labels of the training set by assigning pseudo-labels to unlabeled nodes with high confidence, enabling the re-training of a supervised model with given labels and pseudo labels.

Typically,~\citet{li2018deeper} first chooses top-K high-confidence unlabeled nodes in a self-trained GCN to enlarge the training set for model re-training. 
Both ~\citet{sun2020multi} and ~\citet{hu2021rectifying} propose to generate pseudo-labels via feature clustering in a multi-stage learning schema.
DSGCN~\cite{zhou2019dynamic} selects unlabeled nodes with prediction probabilities higher than a threshold and assigns soft label confidence to
them as label weight.
RS-GNN~\cite{dai2022towards} proposes to use label smoothing regularization for assigning pseudo-labels to nodes based on the edge weight. The underlying intuition is that nodes connected by edges with higher weights are more likely to have the same label.
Different from the above selection-based methods according to certain criteria or generation-based methods according to graph characteristics,
InfoGNN~\cite{li2023informative} significantly identifies another question of how to effectively learn from generated pseudo labels (which might be unreliable) with ground-truth labels given.
Hence, InfoGNN pseudo-labels the most informative nodes through mutual information maximization to alleviate information redundancy and develops a generalized surrogate optimization objective to address noisy and unreliable pseudo-labels.

In summary, graph pseudo-labeling is a straightforward solution to address the scarcity of labels in graph data.
By assigning pseudo-labels to unlabeled nodes in selection-based or generation-based strategies, the graph label set can be augmented to better supervise the training of graph machine learning models.

\subsubsection{Graph Label Denoising} Noisy labels of real-world graphs could significantly damage the effectiveness of the developed GML models, since noisy or mistaken information would propagate along graph structure, and noisy or false labels of nodes will negatively affect their neighbors~\cite{dai2021nrgnn}.  
The core challenges of graph label denoising lie in two folds: (1) how to identify the noisy labels; and (2) how to correct the noisy labels or alleviate the negative effects caused by noisy labels.

Typically, CLNode~\cite{wei2023clnode} focuses on the first challenge and proposes a global difficulty measurement to identify mislabeled difficult nodes based on node feature analysis. Then, a curriculum learning-based training strategy is developed to incrementally involve training nodes for learning expressive GNNs.
In contrast, D-GNN~\cite{nt2019learning} attempts to reduce the effects of noisy labels on GNNs in the weakly supervised learning way, and applies the label noise estimation~\cite{patrini2017making} and loss correction techniques~\cite{patrini2016loss} to develop a surrogate optimization objective with noise resistance.
Besides, IFC-GCN~\cite{hu2021rectifying} first generates the pseudo-labels via feature clustering and adopts an Expectation-Maximization (EM)-like framework to iteratively rectify the pseudo labels and update the node features.
NRGNN~\cite{dai2021nrgnn} proposes to correct the unlabeled/noisy nodes by linking them with correctly-labeled nodes based on the feature similarity criterion. 
The intuition behind this is that if two nodes have high feature similarity, they are more likely to have the same label.
Different from NRGNN which only utilizes label supervision, RTGNN~\cite{qian2023robust} introduces self-reinforcement and consistency regularization as supplemental supervisions, and discriminates the clean and noisy labels in the training process based on the small-loss criterion~\cite{han2018co}.

In summary, addressing noisy labels in graph-structured data is crucial for improving the reliability and performance of graph machine learning models, and these different approaches offer different strategies, for example, designing certain metrics to identify noisy labels (\ie, metric-based) or deriving surrogate objective functions to alleviate negative effects caused by noisy labels (\ie, surrogate loss-based), such that the noisy labels on graph data can be well identified and corrected for benefiting graph machine learning model training and real-world graph applications.

\subsubsection{Graph Class-imbalanced Sampling}
Class-imbalanced graph-structured data refers to the graphs with  biased node-level or graph-level label distributions, where the number of examples for each class may differ significantly~\cite{park2021graphens,zhou2023graphsr}. 
A certain class with a large number of samples is the majority class, whereas a different class with only a small number of samples is the minority class.
When learning on class-imbalanced graph data, graph machine learning models may exhibit bias towards the majority classes, disregarding the minority classes. This can result in overfitting to majority-class examples and consequently lead to reduced performance on minority-class examples. 
This presents a challenging graph-centric issue in developing effective and unbiased GML models due to the class imbalance in label distributions on graphs. Concretely, based on the generic class-imbalanced learning strategy~\cite{he2009learning,chawla2002smote,lin2017focal,cui2019class}, imbalanced classification methods on graphs could also be broadly divided into two categories: graph class-imbalanced sampling methods and cost-sensitive graph learning methods. 
This survey primarily concentrates on graph class-imbalanced sampling methods, which prioritize the perspective of graph data-centric approaches.

Typically, GraphSMOTE~\cite{{zhao2021graphsmote}} extends the generic data resampling method, \ie, SMOTE~\cite{chawla2002smote}, to the 
node embedding spaces on graphs with the synthetic minority oversampling algorithms to synthesize new examples.
And ImGAGN~\cite{qu2021imgagn} also adopts the synthetic oversampling algorithm to only operate on the minority class with a generator.
Moreover, PC-GNN~\cite{liu2021pcgnn} samples nodes and edges with a label-balanced neighborhood sampler to construct balanced sub-graphs for node classification with GNNs. 
In contrast, Graph Mixup~\cite{wang2021mixup}, GraphENS~\cite{park2021graphens}, and GraphSR~\cite{zhou2023graphsr} mix or synthesize features and edges for minority node or other unlabelled nodes to enrich and supplements the minority classes to improve
the class-imbalanced node classification on graphs.
Different from the above-mentioned general class-level imbalance only considering label distributions, ReNode~\cite{chen2021topology} utilizes both the topological structure of the graph and the class distribution information with the Topology-Imbalance Learning (TIL) method. 
TopoImb~\cite{zhao2022topoimb} further addresses the more fine-grained sub-class imbalance on graphs from the perspective of the structural topology imbalance with a topology extractor.
Besides the above node-level methods, igBoost~\cite{pan2013graph} and G2GNN~\cite{wang2022imbalanced} sample and construct balanced sub-graphs respectively, for dealing with the graph-level class-imbalanced issue for classification. 

In summary, effectively addressing the class imbalance problem in graph-structured data from multiple perspectives, including node level, structure level, and graph level, is essential for the development of effective and unbiased graph machine learning models. By employing techniques such as synthesizing minority classes and downsampling majority classes, it is possible to obtain class-balanced graph-structured data in a graph data-centric way, thereby benefiting the training of graph machine learning models.
\begin{table}[t]
\caption{Summary of graph data-centric size enhancement methods.}
\label{tab:size}
\centering
\resizebox{\textwidth}{!}{
\begin{tabular}{p{4.5cm}p{3.5cm}ll}
\toprule
\begin{tabular}[c]{@{}l@{}}Graph Data-centric \\ Size Enhancement\end{tabular} & Methods                              & Techniques                                      & Categories         \\\midrule
\multirow{9}{*}{Graph Size Reduction}                                          
& Herding~\cite{welling2009herding}    & Whole-graph node cluster center based selection & Graph Coreset      \\
& Random~\cite{rebuffi2017icarl}       & Random selection                                & Graph Coreset      \\
& K-Center~\cite{sener2018active}      & K-nearest node cluster center based selection   & Graph Coreset      \\
& Jin et al.~\cite{jin2020gcoarse}     & Spectral property preservation                  & Graph Coarsening   \\
& Cai et al.~\cite{cai2021graph}       & Grouping super nodes                            & Graph Coarsening   \\
& GCOND~\cite{jin2021Gcond}            & Online gradient matching                        & Graph Condensation \\
& DosCond~\cite{jin2022DosCond}        & One-step gradient matching                      & Graph Condensation \\
& CDC~\cite{ding2022faster}            & Distribution preservation                       & Graph Condensation \\
& SFGC~\cite{zheng2023structure}       & Structure-free parameter matching               & Graph Condensation \\
\midrule
\multirow{11}{*}{Graph Data Augmentation}                                      
& DropEdge~\cite{rongdropedge}         & Randomly remove edges                           & Perturbation-based        \\
& GRAND~\cite{feng2020graph}           & Stochastically discard nodes                     & Perturbation-based        \\
& NASA~\cite{bo2022regularizing}       & Random neighbor replacement strategy            & Perturbation-based        \\
& NodeAug~\cite{wang2020nodeaug}       & Graph properties based probability              & Perturbation-based     \\
& GAUG~\cite{zhao2021data}             & Learnable edge sampling probability             & Perturbation-based         \\
& GraphMix~\cite{verma2021graphmix}    & Hidden states and labels interpolation          & Synthetic sample-based        \\
& GraphMixup~\cite{wang2021mixup}      & Node/Edge embedding mixup                       & Synthetic sample-based        \\
& GraphSMOTE~\cite{zhao2021graphsmote} & Node embedding based synthesis                  & Synthetic sample-based        \\
& FLAG~\cite{kong2022robust}           & Gradient-based adversarial perturbations        & Perturbation-based         \\
& G-Mixup~\cite{han2022g}              & Graphon-based generator interpolation           & Synthetic sample-based        \\
& LAGNN~\cite{liu2022local}            & Conditional distribution based synthesis        & Synthetic sample-based \\
\bottomrule      
\end{tabular}
}
\end{table}
\begin{figure}[t]
    \centering\includegraphics[width=0.75\textwidth]{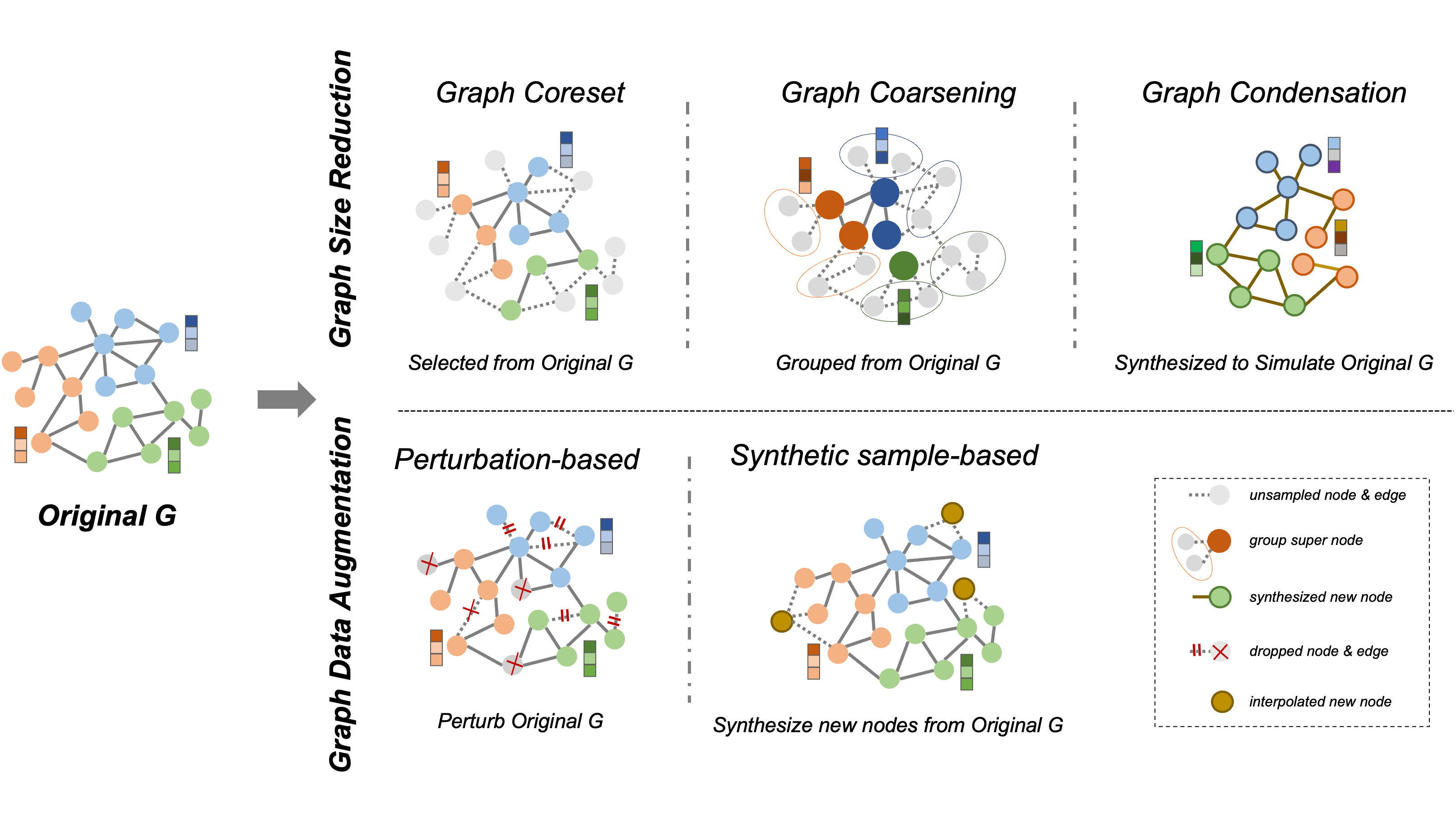}\caption{Illustration of graph data-centric size enhancement methods.}
    \label{fig:Gsize}
\end{figure}

\subsection{Graph Size Enhancement}~\label{subsec:size}
For data-centric graph machine learning, the quantity and scale of input graph data is a crucial factor in affecting the design choice and performance of graph learning models.
Within the different sizes and scales of graph data, we consider two scenarios: oversized large-scale graphs with redundant information and small-scale graphs with limited data sources and insufficient information.
These two scenarios correspond to two types of solutions, \ie, graph size reduction methods and graph data augmentation methods.
An overview of typical techniques used in graph data-centric size-based tasks is given in Fig.~\ref{fig:Gsize}, and a  summary of the methods in graph data-centric size-based tasks is listed in Table.~\ref{tab:size}.

\subsubsection{Graph Size Reduction}
Oversized large-scale graphs might involve noisy or messy nodes and edges, and such redundant information on graph data would significantly affect the performance of designed models.
Moreover, modeling such large-quantity graph data would incur heavy computation and storage costs, damaging the efficiency of graph learning models.
To address the above challenges, graph size reduction methods are proposed to reduce the number of nodes and edges in large-scale graphs to the smaller counterparts, meanwhile, different information from the original graphs is specifically kept in the reduced graphs according to different criteria. 
Typically, graph size reduction methods contain the following techniques, \ie, graph coreset, graph coarsening, and graph condensation, which are shown in Fig.~\ref{fig:Gsize}.

\vspace{1mm}\noindent
\textit{Graph Coreset} identifies the most informative nodes from an original graph and uses those nodes and corresponding edges to build a smaller subgraph as the representative substitution of the original graph. 
The key to graph coreset is the specific criterion, which is used to determine which nodes are considered informative and should be selected.
Typically, Herding method~\cite{welling2009herding,rebuffi2017icarl} minimizes the distance between coreset nodes and whole-graph node cluster centers, while K-Center~\cite{sener2018active} selects the center nodes to minimize the largest distance between a node and its nearest node cluster center.
When no specific criterion is used, the nodes are randomly selected~\cite{welling2009herding}, which is the random graph sampling method.
There is one limitation of graph coreset methods, \ie, their performance is upper-bounded by the information in the selected nodes and edges.
That means, if no inherently representative samples exist in the original graph, the derived subgraphs would fail to achieve optimal performance.

\vspace{1mm}\noindent
\textit{Graph Coarsening} merges and groups nodes and their associated edges in the original graph to `super' nodes with reconnected edges for building a coarser graph with reduced size. 
Certain criteria, such as node connectivity or graph spectrum, are also utilized to guide the coarsening process.
Typically,~\citet{cai2021graph} learned the connections of grouped super nodes in a GNN framework to improve the graph coarsening quality, and~\citet{jin2020gcoarse} maintained the spectral properties of the original graph in the reduced coarser graph, so that valuable information of the graph structure and context can be preserved.
Some works~\cite{buffelli2022sizeshiftreg,huang2021scaling} take graph coarsening as an important pre-processing step to improve the scalability and generalization ability of graph machine learning models, enabling its wide-range applications in large-scale graph understanding and modeling.
The main limitation of graph coarsening is consistent with graph coreset, when the information of grouped nodes is still up-bounded by the original graph. 

\vspace{1mm}\noindent
\textit{Graph Condensation} directly synthesizes a small-scale graph by minimizing the performance gap between GNN models trained on the synthetic, simplified graph and the original oversized graph~\cite{jin2021Gcond}.
That means the only criterion for condensing graphs is the test performance consistency of GNNs.
Typically, GCOND~\cite{jin2021Gcond} first proposes graph condensation with gradient matching between GNNs trained on the original large-scale graph and the synthetic graph, so that the derived small-scale graph could mimic the GNN training behavior of the large-scale graph under the bi-level optimization framework. 
DosCond~\cite{jin2022DosCond} further proposes single-step gradient matching to synthesize graph nodes with a simplified optimization objective and adapts it to the graph classification task.
CDC~\cite{ding2022faster} preserves the distribution of the large-scale graph to condense it to a smaller size for the hyperparameter optimization task in GNNs for serving efficient graph learning.
And SFGC~\cite{zheng2023structure} proposes a structure-free graph condensation paradigm to condense large-scale graphs into a small-scale node set, \ie, graph-free data.
Despite the promising performance of reducing the graph size and approximating the original test performance, current graph condensation performance highly relies on the imitation accuracy of the GNN training behavior. 
This makes existing graph condensation methods poor transferability of condensed graphs for different GNN architectures.

\subsubsection{Graph Data Augmentation}

For the limited data source scenario, small-scale graphs with insufficient information lack adequate training data to instruct the process of graph data learning, leading to the under-fitting issue of models.
In this case, graph augmentation methods are developed to exploit or generate more graph data with adequate quantity~\cite{liu2023data}. 
According to their specific operations to create augmented samples, we divide the graph data augmentation methods into two types: perturbation-based methods and synthetic sample-based methods.

Inspire by the stochastic transformation-based augmentation for image data~\cite{chen2020simple}, the perturbation-based graph augmentation methods modify the features and structure of the original graph to create augmented samples. To conduct perturbation, a simple strategy is to modify the original graph data randomly. Typically, DropEdge~\cite{rongdropedge} is a representative method that proposes to randomly remove a fraction of edges at each training epoch. This simple augmentation strategy is proven to be useful for alleviating over-smoothing and over-fitting problems in GNNs. In GRAND~\cite{feng2020graph}, a similar operation termed DropNode is proposed to create augmented graphs where a number of nodes are stochastically discarded. 
NASA~\cite{bo2022regularizing} utilizes a random neighbor replacement strategy for graph augmentation. In specific, given a target node, some neighboring nodes are removed immediately and then replaced by other close (2-hop or 3-hop) neighbors. 
Different from the above purely stochastic methods, some augmentation approaches use deterministic or learnable probability to control the random modification, which can provide more reliable augmented samples for graph learning. For instance, NodeAug~\cite{wang2020nodeaug} executes feature-level and edge-level random modifications where probabilities are defined by data properties such as node degree. GAUG~\cite{zhao2021data} proposes two advanced strategies for edge modifications: using a pre-trained graph autoencoder to provide the edge sampling probability or jointly optimizing the edge sampling probability with the training of backbone GNN models. 
Besides, adversarial training is also a promising solution for graph perturbation. As a representative approach, FLAG~\cite{kong2022robust} iteratively modifies node features with gradient-based adversarial perturbations during the training phase.

Different from stochastic perturbation-based approaches that apply modifications to the original data, synthetic sample-based graph data augmentation approaches generate synthetic data onto the original data. 
In recent works, Mixup~\cite{zhangmixup} is a commonly used technique to create synthetic images by interpolating two training samples at feature and label space respectively. Following the idea of Mixup, a line of works uses similar operations for synthetic graph data generalization. Considering the discrete structure and inter-node reliance of graph data, how to interpolate two nodes or two graphs together is an open problem. To answer this question, GraphMix~\cite{verma2021graphmix} proposes to interpolate the hidden states and labels with a fully-connected network. Following this idea, Graph Mixup~\cite{wang2021mixup} mixes the hidden representations of each layer with two-branch graph convolution for node-level tasks. Similarly, for graph-level tasks, two graph representations can also be integrated together. Instead of mixing the representations, G-Mixup~\cite{han2022g} augments graphs by interpolating the graphon-based generator of graphs belonging to different classes. Apart from following the idea of Mixup, GraphSMOTE~\cite{zhao2021graphsmote} borrows the idea from SMOTE~\cite{chawla2002smote} to address the imbalanced node classification problem. Specifically, the synthetic nodes belonging to minority classes are created by combining the features of an anchor node and its nearest neighbor. Then, an inner production-based edge generator is used to create synthetic connections for synthetic nodes. To extend the local neighbors via data augmentation, LAGNN~\cite{liu2022local} leverages a conditional variational autoencoder to model the conditional distribution of features of adjacent neighbors, and then samples synthetic neighboring features from the learned distribution. 
\section{How To Learn From Graph Data With Limited-availability and Low-quality}\label{sec:paradigm}
Despite continuous efforts to enhance graph data, the inherent complexity and diversity in such data may continue to present challenges for practical graph learning tasks. 
In certain situations, directly enhancing or refining the graph data may not be feasible and the obtained graph data that experiences improvement may still be unreliable in instructing the graph model development in the DC-GML framework.
In this case, learning from graph data with limited-availability and low-quality is another critical aspect of graph data exploitation in data-centric graph machine learning.
The graph data exploitation strategies can be taken as a bridge connecting current research on graph machine learning models towards the research centered on graph data engineering.

In this article, we mainly consider four types of strategies and methods of graph data exploitation in DC-GML, with varied degrees of limited availability and low quality of graph-structured data:
(1) \textit{graph self-supervised learning} that extracts the supervision signals from the graph data itself automatically instead of manually annotation given the limited availability of graph data labels. 
(2) \textit{graph semi-supervised learning} that learns from vast amounts of graph data given only a small portion of labeled graph data with limited availability.
(3) \textit{graph active learning} that dynamically selects appropriate graph data to annotate during the training procedure given limited availability and low quality of graph data labels.
(4) \textit{graph transfer learning} that addresses graph data distribution shifts to guarantee better model generalization ability.
\begin{table}[t]
\caption{Summary of methods in graph self-supervised learning.}
\label{tab:self-super}
\centering
\resizebox{\textwidth}{!}{
\begin{tabular}{p{5cm}p{9cm}l}
\toprule
\begin{tabular}[c]{@{}l@{}}Graph Self-supervised Learning\end{tabular} & Techniques                                         & Categories     \\\midrule
GAE~\cite{kipf2016variational}                                                                 & Node dot-product for structure reconstruction      & Generative     \\
MGAE~\cite{wang2017mgae}                                                                       & Regression-based feature reconstruction            & Generative     \\
ARGA~\cite{pan2018adversarially}                                                               & Adversarial training for structure reconstruction  & Generative     \\
DGI~\cite{velickovic2019deep}                                                                  & GNN encoder-based mutual information maximization  & Contrastive    \\
MVGRL~\cite{hassani2020contrastive}                                                            & Graph diffusion with contrastive learning          & Contrastive    \\
GraphCL~\cite{you2020graph}                                                                    & Positive samples augmentation                      & Contrastive    \\
PairwiseDistance~\cite{jin2020self}                                                            & Geometric distance based pseudo labels             & Discriminative \\
NodeProperty~\cite{jin2020self}                                                                & Node-level structural property based pseudo labels & Discriminative \\
GROVER~\cite{rong2020self}                                                                     & Domain knowledge-based supervision                 & Discriminative \\
GMI~\cite{peng2020graph}                                                                       & Node-neighbor representation agreement             & Hybrid         \\
You et al.~\cite{you2020does}                                                                  & Node clustering and graph partitioning             & Discriminative \\
GCA~\cite{zhu2021graph}                                                                        & Adaptive positive samples augmentation             & Contrastive    \\
JOAO~\cite{you2021graph}                                                                       & Automated graph data augmentation                  & Contrastive    \\
AD-GCL~\cite{suresh2021adversarial}                                                     & Adversarial training based graph augmentation      & Contrastive    \\
GraphMAE~\cite{hou2022graphmae}                                                              & Re-masking-based feature reconstruction            & Generative     \\
S$^2$GRL~\cite{peng2022new}                                                                    & Geometric distance based pseudo labels             & Discriminative \\
AutoSSL~\cite{jinautomated}                                                                    & Pseudo-homophily metric pretext evaluation         & Hybrid     \\
\bottomrule   
\end{tabular}
}
\end{table}
\subsection{Graph Self-supervised Learning}\label{subsec:self-supervis}
In conventional learning paradigms, the label information is significant in providing supervision signals for model training. 
However, acquiring label information usually requires prohibitive manual annotation, bringing additional costs to data collection~\cite{liu2021self}. 
Compared to labeled data, unlabeled data is easier to obtain from diverse data sources. 
Actually, unlabeled data also contain rich knowledge that can potentially benefit model training; however, it is hard to directly leverage unlabeled data in the supervised learning paradigm. 
To fill the gap, \textit{self-supervised learning} is a promising learning paradigm that exploits unlabeled data for model training~\cite{liu2022graph,liu2021self,jing2020self}.
The key idea of self-supervised learning is to extract the supervision signals from the data itself automatically instead of manual labels. 
To this end, a series of handcrafted auxiliary tasks (\ie, pretext tasks) is designed as surrogate tasks for models to solve. 
In this case, the self-supervised learning paradigm reduces the engineering effort to create and maintain data.

For graph-structured data, self-supervised learning is a significant learning paradigm for unsupervised graph representation learning and model pre-training~\cite{liu2022graph}. 
In recent years, various pretext tasks have been designed by researchers to effectively extract supervision signals from graph data and train graph deep learning models.
According to the learning targets of pretext tasks, graph self-supervised learning methods can be divided into four categories: generative, contrastive, discriminative, and hybrid. 

Generative graph self-supervised learning methods take graph reconstruction as the pretext task. 
In specific, in these methods, a generative decoder is employed to reconstruct the graph structure or node/edge features. 
A reconstruction loss that tries to minimize the similarity between reconstructed and original graph data is utilized for model training. 
As a representative structure generative method, GAE~\cite{kipf2016variational} calculates the dot-product of the representations of two nodes to rebuild the edges in the original graph structure. A binary cross-entropy loss is used as the edge reconstruction loss. 
ARGA~\cite{pan2018adversarially} further introduces an adversarial training strategy to regularize the structure reconstruction model. 
For feature generation, MGAE~\cite{wang2017mgae} regards recovering raw features from noisy input features as the pretext task. 
A regression-based loss function, mean squared error, is utilized to train the MGAE model. 
Hou et al.~\cite{hou2022graphmae} propose GraphMAE that further adds a re-masking mechanism to the decoder input for feature reconstruction. In GraphMAE, a scaled cosine error is used to train the self-supervised learning model.

Contrastive graph self-supervised learning works usually design the pretext tasks following the principle of mutual information maximization. 
That is, maximizing the similarities between the representations of samples with shared semantic information (\ie, positive samples) while minimizing the similarities between the representations of samples with irrelevant semantic information (\ie, negative samples). 
In contrastive methods, data augmentation is a commonly used strategy to construct sufficient positive and negative samples. After the graph learning model generates the representations, a contrastive loss function (\eg, InfoNCE~\cite{tian2020contrastive}) aims to regularize the representation-level similarities of corresponding positive and negative samples. 
Among the contrastive methods, DGI~\cite{velickovic2019deep} is a pioneering one that trains the GNN encoder by maximizing the agreement between each node and the whole graph. To prevent the model from collapsing, DGI introduces full graph shuffling to create negative samples. 
Following the framework of DGI, MVGRL~\cite{hassani2020contrastive} further introduces the graph diffusion technique to create another view for contrastive learning. 
Apart from the node-graph Infomax framework in DGI, SimCLR-like contrastive learning framework~\cite{chen2020simple} is also widely used in contrastive graph self-supervised learning. For example, GraphCL~\cite{you2020graph} introduces four types of graph augmentation, \ie, node dropping, edge perturbation, attribute masking, and subgraph extraction, to generate augmented positive samples. Then, an InfoNCE~\cite{tian2020contrastive} is leveraged to maximize the agreement between two augmented samples of the same anchor graph. 
GCA~\cite{zhu2021graph} applies a similar framework to node-level graph representation learning tasks and further employs adaptive augmentation strategies to create high-quality positive samples. 
JOAO~\cite{you2021graph} utilizes a bi-level optimization framework to select suitable augmentation strategies for different graph datasets automatically. 
To avoid capturing redundant information during graph contrastive learning, Suresh et al.~\cite{suresh2021adversarial} propose to use adversarial training for graph augmentation. 

Discriminative graph self-supervised learning methods capitalize pseudo labels or auxiliary properties to enrich the supervision signals for model training. With the surrogate labels from auxiliary information, the graph learning models can be trained in a supervised manner. For instance, Node Clustering and Graph Partitioning~\cite{you2020does} create pseudo labels via feature-based clustering algorithm and structure-based graph partitioning algorithm respectively. Then, the GNN encoder can be optimized with a supervised classification loss (\ie, cross-entropy) using the pseudo labels. S$^2$GRL~\cite{peng2022new} and PairwiseDistance~\cite{jin2020self} extract the geometric distance between two nodes as the pseudo labels of the corresponding node pair. NodeProperty~\cite{jin2020self} extracts node-level structural properties (\eg, node degree) as continuous labels to guide regression-based model training. GROVER~\cite{rong2020self} introduces domain knowledge-based supervision signals for discriminative self-supervised learning tasks. Concretely, the existence of key motifs in molecules serves as the pseudo labels to pre-train the Transformer-based graph learning model for molecular data.

Hybrid graph self-supervised learning methods aim to provide better self-supervision by integrating multiple pretext tasks in a joint learning manner. Typically, GMI~\cite{peng2020graph} a contrastive pretext task is employed to maximize the agreement between node representations and its neighbors' features; at the same time, a generative pretest task is built to reconstruct the graph adjacency matrix. 
To balance the contributions of multiple pretext tasks in hybrid scenarios, AutoSSL~\cite{jinautomated} utilizes `pseudo-homophily' as a metric for pretext task quality and uses the evolution algorithm and meta-gradient to automatically optimize the allocation of self-supervised learning tasks in a joint learning model.
\begin{table}[t]
\caption{Summary of methods in graph semi-supervised learning.}
\label{tab:semi-super}
\centering
\resizebox{\textwidth}{!}{
\begin{tabular}{p{5.5cm}p{8.5cm}l}
\toprule
\begin{tabular}[c]{@{}l@{}}Graph Semi-supervised Learning\end{tabular} & Techniques                                            & Categories           \\\midrule
Zhu et al.~\cite{zhu2003semi}                                                                  & Graph Laplacian regularization                        & Regularization-based \\
Zhou et al.~\cite{zhou2003learning}                                                            & Graph Laplacian regularization                        & Regularization-based \\
Zhou et al.~\cite{zhou2005learning}                                                            & Local smoothness under homophily                      & Regularization-based \\
Li et al.~\cite{li2018deeper}                                                                  & Self-training with training set extension             & Pseudo-labelling   \\
NodeAug~\cite{wang2020nodeaug}                                                                 & KL divergence-based consistency                       & Regularization-based \\
GRAND~\cite{feng2020graph}                                                                     & L2 distance-based consistency                         & Regularization-based \\
M3S~\cite{sun2020multi}                                                                        & Clustering-based pseudo label generation              & Pseudo-labelling     \\
SimP-GCN~\cite{jin2021node}                                                                  & Feature-level similarity in pairwise distance         & Regularization-based \\
GCN-LPA~\cite{wang2021combining}                                                           & Edge weights with graph structure regularization      & Regularization-based \\
CG$^3$~\cite{wan2021contrastive}                                                               & Self-supervised objective based regularization        & Regularization-based \\
GCPN~\cite{wan2021contrastive_cgpn}                                                            & Contrastive and possion learning based regularization & Regularization-based \\
Meta-PN~\cite{ding2022meta}                                                                    & Adaptive label propagator based on label propagation  & Pseudo-labelling     \\
CycProp~\cite{li2022cyclic}                                                                    & High-quality contextual node selection                & Pseudo-labelling    \\
\bottomrule
\end{tabular}
}
\end{table}
\subsection{Graph Semi-supervised Learning}~\label{subsec:semi-supervis}

Since obtaining large amounts of labeled data can be time-consuming and expensive, how to better leverage both labeled and unlabeled data is a key challenge in data-centric AI.  To this end, semi-supervised learning is a promising paradigm that allows models to learn from vast amounts of data, even when only a small portion of that data is labeled, which can benefit data-centric machine learning. Due to the dependency among node samples, in graph machine learning tasks especially node-level tasks, the majority of methods~\cite{kipf2017semi,velivckovic2018graph,wu2019simplifying} focus on a semi-supervised learning paradigm where a small fraction of labeled nodes and a large number of unlabeled nodes are available for model training. However, most of them only use unlabeled nodes to calculate the representations of labeled nodes during aggregation rather than further mining the interior knowledge from the unlabeled data~\cite{velivckovic2018graph,wu2019simplifying}. Going back to the original intention of semi-supervised learning and its contribution to data-centric AI, in this subsection, we mainly focus on the graph machine learning methods that learn to fully leverage the unlabeled data with specific designs, particularly those for scenarios where labeled samples are extremely few. In concrete, we summarize the semi-supervised methods that follow two pathways to exploit the unlabeled samples, \ie, applying extra regularization to unlabeled data and 
generating pseudo labels for unlabeled data. 

A simple yet effective strategy to leverage unlabeled samples is forcing the model to satisfy several criteria on unlabeled data with regularization. Based on the homophily assumption (\ie, connected nodes are likely to share the same properties), preserving the local smoothness of the learned representations is a natural idea for regularization. 
An early work~\cite{zhou2005learning} proposes a regularization term that sums the weighted variation of each edge in a directed graph. 
In this way, the predictions of two nodes become similar if they are adjacent or densely connected. 
Following the homophily assumption, graph Laplacian regularization~\cite{zhu2003semi,zhou2003learning} is also a commonly used form of explicit regularization for local smoothing. 
Beyond the homophily assumption, Jin et al.~\cite{jin2021node} propose to preserve the node feature-level similarity on the learned representations with a pairwise distance-based regularization. 
Wang et al.~\cite{wang2021combining} propose a label propagation algorithm-based loss function to regularize the edge weights in the adjacency matrix of semi-supervised learning models. Apart from the regularization of smoothness and similarity, consistency training is also a useful semi-supervised learning strategy that regularizes the model to make the predictions consistent with respect to noise injected by data augmentation~\cite{xie2020unsupervised}. Based on the idea of consistency training, NodeAug~\cite{wang2020nodeaug} applies augmentation to each node and uses a KL divergence-based consistency loss to minimize the difference between the predicted distributions of original nodes and augmented nodes. Following a similar idea, GRAND~\cite{feng2020graph} utilizes random node dropping for graph data augmentation and then uses an L2 distance-based consistency loss to regularize the categorical distribution. Considering their powerful ability to leverage unlabeled data, some self-supervised learning objectives can also serve as regularization terms in semi-supervised learning scenarios. For example, CG$^3$~\cite{wan2021contrastive} uses both generative and contrastive self-supervised learning targets to train the semi-supervised GNN model with labeled and unlabeled data. GCPN~\cite{wan2021contrastive_cgpn} integrates contrastive learning and possion learning into a unified framework for node classification with extremely limited labeled data. 


Considering the sparsity of labeled samples, an intuitive solution for graph semi-supervised learning is creating reliable pseudo labels for unlabeled data to extend the training set. To generate pseudo labels with the assistance of graph structure, label propagation, which propagates label information along with edges, is a promising algorithm for pseudo labeling~\cite{wang2006label}. For instance, Meta-PN~\cite{ding2022meta} constructs an adaptive label propagator based on label propagation to generate pseudo labels for nodes with high prediction confidence. 
Similarly, CycProp~\cite{li2022cyclic} uses label propagation to select high-quality contextual nodes for GNN model training. Different from label propagation, clustering algorithms are also effective in generating pseudo labels. 
M3S~\cite{sun2020multi} is a representative clustering-based method that employs DeepCluster technique~\cite{caron2018deep} to provide pseudo labels to optimize GNN models. 
Using the self-training strategy, Li et al.~\cite{li2018deeper} propose to extend the training set using the samples with high prediction confidence. 
\begin{table}[t]
\caption{Summary of methods in graph active learning.}
\label{tab:active}
\centering
\resizebox{\textwidth}{!}{
\begin{tabular}{p{5.5cm}p{9cm}l}
\toprule
\begin{tabular}[c]{@{}l@{}}Graph Active Learning \end{tabular} & Techniques                                                            & Categories       \\\midrule
AGE~\cite{cai2017active}                                        & Information entropy, density, and centrality rules & Rule-based       \\
ANRMAB~\cite{gao2018active}                                     & Multi-armed bandit mechanism                                          & Rule-based       \\
ActiveHNE~\cite{chen2019activehne}                              & Multi-armed bandit mechanism on heterogeneous graphs                  & Rule-based       \\
FeatProp~\cite{wu2019active}                                    & Closest cluster center based labelling                                & Clustering-based \\
ATNE~\cite{jin2020active}                                       & Active transfer learning based node selection                         & Rule-based       \\
ASGN~\cite{hao2020asgn}                                         & Sample diversity based node selection                                 & Rule-based       \\
GPA~\cite{hu2020graph}                                          & GCN-based policy network                                              & RL-based         \\
MetAL~\cite{madhawa2020metal}                                   & Meta-gradients estimation                                             & Meta Learning-based     \\
SEAL~\cite{li2020seal}                                          & Adversarial learning with divergence value                            & Adversarial-based    \\
GRAIN~\cite{zhang2021grain}                                     & Diversified influence maximization objective                          & Influence-based  \\
RIM~\cite{zhang2021rim}                                         & Label reliability based influence score scaling                       & Influence-based  \\
Attent~\cite{zhou2021attent}                                    & Active graph alignment                                                & Influence-based  \\
ALG~\cite{zhang2021alg}                                         & Clustering-based density \& Attention-based score                     & Metric-based     \\
ALLIE~\cite{cui2022allie}                                       & Integrated graph coarsening and focal loss                            & RL-based         \\
BIGENE~\cite{zhang2022batch}                                    & Q-value decomposition with batch sampling selection                   & RL-based         \\
IGP~\cite{zhanginformation}                                     & Information gain propagation for soft labelling                       & Influence-based  \\
JuryGCN~\cite{kang2022jurygcn}                                  & Jackknife uncertainty estimation                                      & Influence-based  \\
\bottomrule
\end{tabular}
}
\end{table}
\subsection{Graph Active Learning}~\label{subsec:active}

In data-centric AI, the number of training samples is usually the bottleneck of model performance. Considering that the cost (\eg, human labor and expert knowledge) for label annotation is usually fixed, how can we fully make the best use of such a labeling budget? Active learning is a promising learning paradigm to answer this question. The core idea of active learning is to dynamically select the samples to label during the training procedure. In the practical active learning process, the nodes to label are selected automatically by the models following several selection criteria. In this way, the labeling cost can be minimized and the model performance can be further improved~\cite{gao2018active}. Unlike the independent and identically distributed (i.i.d) data such as images and texts, the samples (\ie, nodes) in graph-structured data are densely interconnected, leaving a significant challenge for \textit{graph active learning}. Therefore, new selection criteria as well as active learning frameworks with the consideration of data dependency should be specially designed. In the following paragraphs, we categorize existing graph active learning methods into several types, including rule-based, reinforcement learning-based, influence function-based, and other advanced methods. 

In graph active learning, the core is how to select appropriate samples to label. Early methods mainly consider several \textit{pre-defined rules} as the selection criteria in graph active learning. The pioneering work AGE~\cite{cai2017active} proposes three basic rules: information entropy, information density, and graph centrality. In AGE, the information entropy measures the prediction of classification output, the information density measures the attribute representative of nodes in a latent space, and graph centrality measures the structural representative of nodes with the consideration of their topological roles. To balance the contribution of different rules, AGE introduces multiple hyper-parameters for trade-offs. 
To automatically make decisions from multiple selection rules, ANRMAB~\cite{gao2018active} employs a multi-armed bandit mechanism to adaptive select the most important nodes for labeling. Following ANRMAB, ActiveHNE~\cite{chen2019activehne} further applies the multi-armed bandit mechanism to active representation learning on heterogeneous graphs. Based on the expectation of certainty, ATNE~\cite{jin2020active} conducts accurate node selection for active transfer learning on the network embedding task. For active learning in molecular property prediction scenarios, ASGN~\cite{hao2020asgn} adopts the sample diversity to measure the importance of samples during selection. 

Although the rule-based graph active learning methods enjoy simple learning procedures, they may ignore the interactions between nodes when making decisions. Moreover, their greedy selection strategies make them sometimes neglect the long-term performance gain of the whole query sequence~\cite{hu2020graph}. To overcome these limitations, an advanced solution is approaching graph active learning via \textit{reinforcement learning}. GPA~\cite{hu2020graph} is one of the representative reinforcement learning-based methods. It utilizes a GCN-based policy network to execute the action of selecting nodes to label, where the input state includes various heuristic criteria (\eg, degree, entropy, and local similarity) and the reward is acquired from validation accuracy. To handle large-scale and imbalance problems in graph active learning, ALLIE~\cite{cui2022allie} integrates graph coarsening and focal loss to the reinforcement learning-based graph active learning framework. 
To enable batch sampling selection in graph active learning, BIGENE~\cite{zhang2022batch} uses a multi-agent reinforcement learning framework with Q-value decomposition to avoid the explosion of combinatorial action space. 

To effectively consider the mutual impact between nodes during active sample selection, another line of approaches uses \textit{influence function} to evaluate the contribution of the selected samples in graph active learning. Based on the social influence theorem~\cite{li2018influence}, the influence-based methods usually use node feature influence functions~\cite{wang2021combining} as the key criterion in graph active learning. For example, GRAIN~\cite{zhang2021grain} introduces a diversified influence maximization objective to optimize active sample selection. RIM~\cite{zhang2021rim} further leverages label reliability to scale the influence-based scores for node selection. To solve the active graph alignment problem, Attent~\cite{zhou2021attent} utilizes the influence function computed from the alignment solution to evaluate the quality of nodes to select. To execute graph active learning in a soft-label setting, IGP~\cite{zhanginformation} introduces a new criterion termed information gain propagation as the maximization objective for node selection. Following the idea of jackknife uncertainty estimation, JuryGCN~\cite{kang2022jurygcn} utilizes the influence functions to estimate the model parameter change without re-training the GNN models, and the estimated uncertainty can be further used for active learning. 

Apart from the aforementioned typical strategies for active sample selection, there are also some works that try to apply various advanced techniques to graph active learning. For instance, FeatProp~\cite{wu2019active} uses distance-based clustering for active learning where the node closest to each cluster center is picked to label. To conduct graph active learning effectively, ALG~\cite{zhang2021alg} considers two cost-effective measurements, \ie, a clustering-based density and an attention-based score, and employs a receipted field maximization strategy to choose candidate nodes. MetAL~\cite{madhawa2020metal} introduces meta-gradients to estimate the impact of candidate nodes on the model performance and then guide the selection. SEAL~\cite{li2020seal} integrates adversarial learning into graph active learning framework. In SEAL, a logit-based divergence value from the adversarial learning model is used to evaluate the informativeness of nodes. 
\begin{table}[t]
\caption{Summary of methods in graph transfer learning.}
\label{tab:transfer}
\centering
\resizebox{\textwidth}{!}{
\begin{tabular}{p{5.5cm}p{9cm}l}
\toprule
Graph Transfer Learning                                 & Techniques                                          & Categories                \\\midrule
DANE~\cite{zhang2019dane}         & Adversarial learning regularization                 & Close-set shift           \\
UDA-GCN~\cite{wu2020unsupervised} & Adversarial learning with dual-GNN                  & Close-set shift           \\
ACDNE~\cite{shenDC0C20}           & Node affinity \& topological proximity preservation & Close-set shift           \\
OpenWGL~\cite{wu2020openwgl}      & Variational graph autoencoder                       & Open-set shift            \\
PGL~\cite{luo2020progressive}     & Class space decomposition                           & Open-set shift            \\
SRGNN~\cite{zhu2021srgnn}         & Central moment discrepancy (CMD) measurement        & Close-set shift           \\
SOGA~\cite{mao2021soga}           & Mutual information maximization                     & Close-set shift           \\
DGDA~\cite{cai2021graphda-view}   & Domain and semantic separation                      & Close-set shift           \\
SRNC~\cite{zhu2022shift}          & Unified domain adaption GNN                         & Close-set/Open-set shifts\\\bottomrule
\end{tabular}
}
\end{table}
\subsection{Graph Transfer Learning}~\label{subsec:transfer}
Graph data always varies along the time and processing stages, when new node classes might emerge with the enlargement of graph data and new edges and connections might happen between the observed graph data and new coming graph data.
For instance, in citation networks, the connections (the edges) for paper citations (the nodes) and subject areas/topics (the classes) would go through significant change as time goes by~\cite{hu2020open}.
Except for focusing on the availability and quality of graph data itself, modeling graph data shifts in cross-domains and various distributions~\cite{guo2023data} is also an important aspect of DC-GML, as the domain and distribution shifts of graph data would significantly affect the generalization ability and robustness of graph machine learning algorithms.
\begin{figure}[t]
    \centering\includegraphics[width=0.8\textwidth]{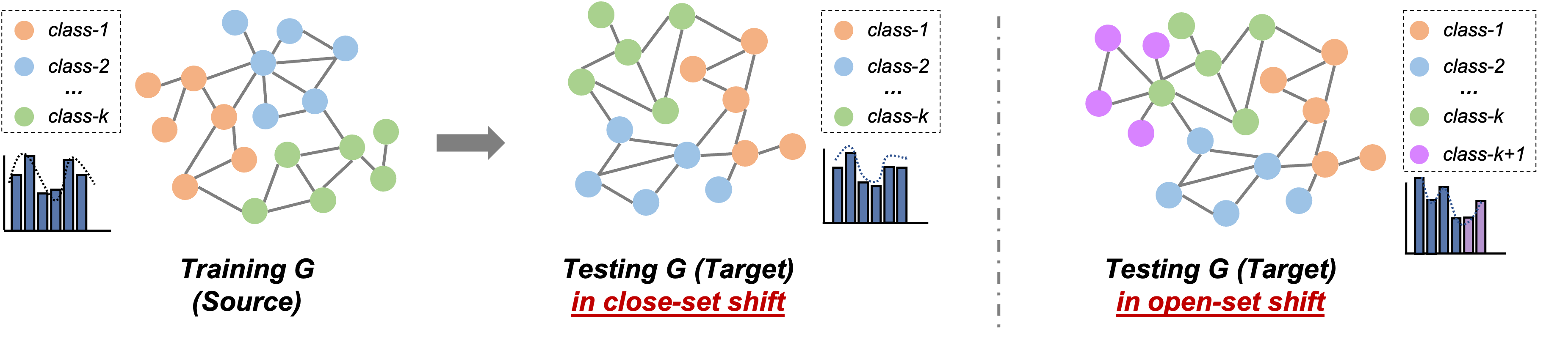}\caption{Illustration of graph transfer learning in graph data-centric close-set shift and open-set shift.}
    \label{fig:Gtrasnfer}
\end{figure}

In light of this, graph transfer learning emerges as a crucial strategy to address the challenge of distribution shifts in graph data over time.
In this way, graph machine learning models trained with observed graph data in a domain under a certain distribution, are expected to perform well for unobserved graph data in other domains under different distributions.
Hence, we center on the distribution-shift graph data in different domains and sources, and comprehensively summarize the data-centric graph transfer learning methods. 
And we mainly focus on the graph domain adaption task, which is a vital aspect of graph transfer learning, specifically addressing the testing and training graph data distribution shift, \ie, not identically and independently distributed (i.i.d.) issue.

According to whether label spaces of graphs are changed or not, Zhu et al.~\cite{zhu2022shift} categorize graph domain adaption methods into two groups: (1) open-set shift~\cite{zhu2022shift,wu2020openwgl,luo2020progressive}, where new classes and label spaces emerge in the target graph domain; and (2) close-set shift~\cite{zhu2022shift,zhang2019dane,wu2020unsupervised,zhu2021srgnn,mao2021soga,cai2021graphda-view,pilanci2020domain,shenDC0C20}, where no new classes emerge but the distributions of features, structures, and the labels all shift in the target domain. The illustration of graph transfer learning in graph data-centric close-set shift and open-set shift is in Fig.~\ref{fig:Gtrasnfer}, and the detailed method summary is present in Table.~\ref{tab:transfer}.

Specifically, for the close-set shift, the core idea is to learn domain invariant representation between the source and target data by minimizing the domain discrepancy measures, \eg, central moment discrepancy (CMD)~\cite{zellinger2017cmd}, Fr{\'e}chet distance (FD)~\cite{dowson1982fd}, maximum mean discrepancy (MMD)~\cite{long2017mmd}.
Typically, SRGNN~\cite{zhu2021srgnn} adopts the CMD measurement and importance sampling for in-distribution non-IID training data.
And SOGA~\cite{mao2021soga} enforces the constraint of measuring maximal mutual information and structural proximity between source and target graph data.
Besides, DANE~\cite{zhang2019dane} utilizes adversarial learning regularization to train a domain discriminator, aligning the embedding output distribution of the source and target and enabling its GCN encoder to learn representations that are not specific to a certain domain.
With a similar adversarial learning regularization pattern~\cite{alam2018domain}, UDA-GCN~\cite{wu2020unsupervised} further proposes a dual-GNN architecture to obtain the unified domain invariant representations. 
Additionally, DGDA~\cite{cai2021graphda-view} separates domain and semantic latent variables, while ACDNE~\cite{shenDC0C20} maintains node attribute affinity and topological proximity. 
These two methods aimed to achieve the same goal: learning domain-invariant node representations over graphs.

Different from the close-set shift with invariant node class label spaces, for the open-set shift, unseen classes of nodes pose severe challenges for graph domain adaption, as such nodes have no labeled samples and may exist in an arbitrary uncertain latent representation space different from those seen classes. 
To model the uncertainty of nodes and improve the robustness of open-world GML models, OpenWGL~\cite{wu2020openwgl} designs a variational graph autoencoder to learn node representation sensitive to unseen class with label loss and class uncertainty loss.
PGL~\cite{luo2020progressive} decomposes the original latent hypothesis spaces into seen-class space and unseen-class space, and adopts sample- and manifold-level open-set risk control strategy with the progressive learning paradigm to learn class-specific representations and close the gap between the source and target distributions. 
In contrast, SRNC~\cite{zhu2022shift} proposes the unified domain adaption GNN framework for multiple types
of distribution shifts under both open-set and close-set shifts. 
Specifically, SRNC~\cite{zhu2022shift} co-trains a shift-robust classifier with a variational cluster, which implemented graph clustering to identify latent classes in target data distribution and approximated the source class conditional distribution, leading to better out-of-distribution generalization on open-set and close-set distribution shifts. 

In summary, graph domain adaption within the graph transfer learning contributes to measuring and closing the gap between source and target domain graph data, and then learning domain invariant or class-specific representations for close-set and open-set distribution shift scenarios, respectively. 
The core idea is to transfer the expressive power of graph machine learning models trained on labeled source domain data to unlabeled target domain data, resulting in improved learning ability, robustness, and generalization capabilities for the target graph domain.
In the context of graph data-centric learning, graph transfer learning-based domain adaption improves the performance of models by fully exploiting and shrinking the data discrepancy between source and target, leading to more accurate predictions and better performance of graph learning models.


\begin{table}[t]
\renewcommand\arraystretch{1.25}
\caption{Summary of phases, methods, tools, and goals for building graph MLOps from graph data-centric view.}
\label{tab:graphmlop}
\centering
\resizebox{\textwidth}{!}{
\begin{tabular}{lll}
\toprule
Phases                                  & Goals                                     & Methods \& Tools              \\\midrule
\multirow{2}{*}{Graph Data Collection}  & Graph Data Crowdsourcing                  & \begin{tabular}[c]{@{}l@{}}Amazon Mechanical Turk~\cite{mturk2005}, \\ Tang et al.~\cite{tang2011semi}, Cao et al.~\cite{cao2021knowledge}\end{tabular}\\\cmidrule(r){2-3}
                                        & Graph Data Synthesis                      & \begin{tabular}[c]{@{}l@{}}SBMs~\cite{snijders1997estimation}, Koller et al.~\cite{koller2009probabilistic}, Ying et al.~\cite{ying2019gnnexplainer},\\ Unsupervised methods~\cite{perozzi2014focused, muller2023graph},\\ Semi-supervised methods~\cite{rozemberczki2021pathfinder, dwivedi2020benchmarking, tsitsulin2022synthetic, palowitch2022graphworld}\end{tabular}                                                \\\midrule
\multirow{2}{*}{Graph Data Exploration} & Graph Data Understanding \& Visualization & \begin{tabular}[c]{@{}l@{}}NetworkX~\cite{networkx}, igraph~\cite{igraph}\\ Neo4j~\cite{neo4j}\end{tabular}                       \\\cmidrule(r){2-3}
                                        & Graph Data Valuation                      & GraphSVX~\cite{duval2021graphsvx}\\\midrule
\multirow{2}{*}{Graph Data Maintenance} & Graph Data Privacy                        & \begin{tabular}[c]{@{}l@{}}TrustworthyGNN~\cite{zhang2022trustworthy}, Zhang et al.~\cite{zhang2010privacy},\\ Liu et al.~\cite{liu2008towards}, Yu et al.~\cite{yu2018privacy}, \\ Mulle et al.~\cite{mulle2015privacy}, PGAS~\cite{zhang2020pgas}, \\ Federatedscope-GNN~\cite{wang2022federatedscope}, Tan et al.~\cite{tan2022federated}\end{tabular} \\\cmidrule(r){2-3}
                                        & Graph Data Security                       & \begin{tabular}[c]{@{}l@{}}Sandhu et al.~\cite{sandhu1994access}, Abidi et al.~\cite{abidi2014implementation},\\ Li et al.~\cite{li2021threat}\end{tabular}\\\midrule
\multicolumn{2}{l}{Graph MLOps}                                                     & \begin{tabular}[c]{@{}l@{}}Kubeflow~\cite{kubeflow}, Amazon SageMaker~\cite{sagemaker}, \\Amazon Neptune~\cite{neptune} \end{tabular}\\\bottomrule
\end{tabular}
}
\end{table}
\section{How To Build Graph MLOps Systems: The Graph Data-centric View}\label{sec:systems}
After obtaining and refining graph data through two critical steps of graph data improvement and exploitation, this section takes into account the three remaining phases: graph data collection, exploration, and maintenance, for providing a broader perspective that encompasses the entire graph data lifecycle.
The ultimate goal is to integrate all graph data-centric components into a systematic and comprehensive graph MLOps workflow from the graph-centric view. 
This integration is expected to foster cooperation with graph model-centric components, resulting in a mutually beneficial scenario in the advancement of graph data engineering and graph model development.

Concretely, for graph data collection, we consider two aspects of crowdsourcing and synthesis, which aim to provide sufficient supervision for graph machine learning model training. 
For graph data exploration, we aim to understand, analyze, and manage numerous and complex graph-structured data explicitly and comprehensively for graph machine learning model design. 
For graph data maintenance, we aim to maintain, update, and integrate in-service graph data under security and privacy management for graph machine learning model deployment.
All these compose the systematic graph MLOps workflow within the entire graph data-centric lifecycle. The summary of all phases, methods, tools, and goals for building graph MLOps from graph data-centric view is listed in Table.~\ref{tab:graphmlop}.

\subsection{Graph Data Crowdsourcing and Synthesis}\label{sec:datacollection}
\subsubsection{Graph Data Crowdsourcing} 
Graph data crowdsourcing aims to annotate or refine the labels of graph data through online platforms or tools (\eg, Amazon Mechanical Turk~\cite{mturk2005}) by crowd workers.
Specific initial guidelines are provided to annotators for performing tasks, such as identifying the class categories of nodes or adding missing edges to a graph.
Two significant challenges typically arise when crowdsourcing graph labeling. The first is that the instructions provided by designers may not be clear, which leads to the need for research into \textit{clarifying labeling instructions}. The second challenge is that, even when instructions are clear, diverse workers may not assign consistent labels, leading to the need for research into \textit{consistent labeling}.

Crowdsourcing graph labeling almost shares the same schema, challenges, and insights as that for image data or text data.
For example, Tang et al.~\cite{tang2011semi} jointly compute the expected true labels given the worker labels and estimated worker accuracy on the MTurk document judgment task.
The techniques of handling worker bias and noise in this work can be generally applied to various graph labeling tasks.
Importantly, Cao et al.~\cite{cao2021knowledge} identify the importance of crowdsourcing and propose three key issues on knowledge-intensive crowdsourcing from the perspectives of tasks, workers, and crowdsourcing processes in a comprehensive survey.
Moreover, they proposed to construct Knowledge Graph (KG) data with crowdsourcing for facilitating the development of KG technology. 
Concretely, they proposed that the construction of KG with crowdsourcing can involve the following stages: (1) entity construction; (2) knowledge mining and filling; and (3) refinement of knowledge graphs.  

In summary, crowdsourcing offers a tremendous opportunity to enhance the creation and annotation of graph data, which, in turn, can facilitate the development and training of graph machine learning models. 
Additionally, tasks such as consistent labeling can improve the quality of graph data creation and labeling right from the initial stages of graph data collection through crowdsourcing.

\subsubsection{Graph Data Synthesis} 

Graph data synthesis refers to the process of generating synthetic graph data that closely resembles real-world graphs while preserving their structural properties and statistical characteristics. It plays a crucial role in graph machine learning research by providing a means to generate large-scale and diverse graph datasets for training and evaluation purposes. Graph data synthesis techniques enable researchers to overcome limitations such as data scarcity, privacy concerns, and the need for labeled data.

The goal of graph data synthesis is to create synthetic graphs with specific inherent structural patterns, node attributes, and edge connections. To this end, a mainstream solution is to use predefined generative models to acquire graph data with several properties. For instance, stochastic block models (SBMs)~\cite{snijders1997estimation} are a family of representative graph generative models that generate edges of graphs conditioned on some predefined partition of a graph. Probabilistic graphical models~\cite{koller2009probabilistic} can also be a group of reasonable graph generative models that represent and reason uncertainty by combining probability theory and graph theory to model dependencies and relationships among random variables. Based on these generative models, advanced algorithms can be developed to generate synthetic graphs for various scenarios, covering unsupervised graph learning~\cite{perozzi2014focused, muller2023graph} and semi-supervised learning~\cite{rozemberczki2021pathfinder, dwivedi2020benchmarking, tsitsulin2022synthetic, palowitch2022graphworld}. Another line of studies generates synthetic graph data via motif-based graph generation. For example, Ying et al.~\cite{ying2019gnnexplainer} combine Barabasi-Albert (BA) graph with different graph motifs (\eg, house and cycle) to construct synthetic datasets for the evaluation of explainable GNNs. Following works~\cite{luo2020parameterized, wu2022discovering} further introduce novel combination strategies and motifs for graph construction. 

In conclusion, graph data synthesis is a vital component of data-centric graph machine learning research. By generating realistic and diverse graph data, researchers can drive innovation, improve algorithmic understanding, and contribute to the broader field of graph analytic. Continued advancements in graph data synthesis techniques will empower us to tackle complex real-world challenges using graph-based approaches.

\subsection{Graph Data Understanding, Visualization, and Valuation}\label{sec:dataexploration}
Real-world graph-structured data is usually in large quantity with massive nodes and edges, showing great diversity and complexity.
In this case, it is usually hard to explicitly and comprehensively understand, analyze, and manage numerous and complex graph structure data.
To facilitate human researchers' better graph data understanding, various graph data visualization tools are designed to transfer high-dimensional and complexly connected graph data to low-dimensional or well-presented visualization representations, so that we can gain insights into the complex interconnections and dependencies among entities, understand how entities are connected, the strength of relationships, and identify patterns or clusters within graphs.

In this paper, we list several commonly-used graph data understanding and visualization tools: \textit{(a) NetworkX}~\cite{networkx}: a Python library for the creation, manipulation, and study of the structure, dynamics, and functions of complex networks. It also provides basic graph visualization capabilities, allowing you to create visual representations of graphs using Matplotlib or other visualization libraries; \textit{(b) igraph}~\cite{igraph}: a popular open-source library for creating, manipulating, and analyzing networks in various programming languages such as R, Python, and C/C++. It offers basic graph visualization capabilities, including customizable plotting functions for visualizing graphs and networks; \textit{(c) Neo4j}~\cite{neo4j}: a graph database management system (DBMS). It is specifically designed to store, manage, and query graph-structured data. It leverages the property graph model, where nodes represent entities, relationships define connections between nodes, and properties hold additional information about nodes or relationships, allowing rich and expressive representations of real-world relationships in graph data.
\textit{(d) Amazon Neptune}~\cite{neptune}: a fully managed graph database service provided by Amazon Web Services (AWS). Neptune supports the property graph model and allows you to run highly available and scalable graph workloads in the cloud.

In addition, graph data valuation is another promising aspect in data-centric graph ML within the context of data valuation, which aims to understand the value of a data point for a given machine learning task. It is an essential primitive in the design of graph data marketplaces and explainable AI~\cite{zha2023dcaisurvey,zha2023dcaisurvey,duval2021graphsvx}.
Concretely, to understand how each data point contributes to the final performance, researchers could develop certain metrics, \eg, the Shapley value, on the data points to estimate their contributions.
For instance, GraphSVX~\cite{duval2021graphsvx} captures the `fair' contribution of each feature and node towards the explained prediction by constructing a
surrogate model on a perturbed graph dataset, so that it could provide explanations according to the Shapley values.


\subsection{Graph Data Privacy and Security}\label{sec:datamaintain}

Graph data privacy and security are significant concerns in the field of data-centric graph machine learning. The increasing accessibility and sharing of graph data present risks of privacy violations and malicious attacks~\cite{zhang2022trustworthy}. In real-world scenarios, graphs often contain sensitive information, necessitating the protection of individual privacy and the integrity of the data. As a result, unique privacy challenges in graph data arise due to the interconnected nature of nodes and edges within a graph~\cite{zhang2010privacy}. In this case, specialized techniques are required to address these challenges.

To tackle the privacy challenges, various privacy-preserving techniques have been proposed for graph data. For instance, anonymization techniques aim to obfuscate the data while maintaining the graph's structure and statistical properties~\cite{liu2008towards}. Meanwhile, perturbation techniques, involving adding noise or modifying connectivity, can also help protect the privacy of individuals and relationships~\cite{yu2018privacy, mulle2015privacy}. Also, graph encryption ensures that only authorized parties with appropriate decryption keys can access and interpret the data~\cite{zhang2020pgas}. Finally, as a booming research direction, federated graph learning enables collaborative model training without directly sharing the raw graph data, minimizing data exposure~\cite{wang2022federatedscope, tan2022federated}.

In addition to privacy, ensuring the security of graph data is crucial for data-centric graph machine learning. To achieve this goal, various access control mechanisms, such as authentication and authorization, restrict data access to authorized users~\cite{sandhu1994access}. On the other hand, data integrity techniques, such as digital signatures and hash functions, are able to verify that the data hasn't been tampered with~\cite{abidi2014implementation}. Moreover, threat detection and mitigation strategies, including intrusion detection systems and network monitoring, help identify and counter potential security threats~\cite{li2021threat}.

Graph data privacy and security are ongoing areas of research, as new threats and challenges continue to emerge with the advancement of graph-based applications. Researchers and practitioners need to stay vigilant and adopt a multi-faceted approach to address privacy and security concerns in graph data, ensuring that individual privacy is protected, sensitive information is safeguarded, and the integrity of the graph data is maintained.

\subsection{Graph MLOps}\label{sec:mlops}
Machine learning operations (MLOps) is a set of standardized processes and technology capabilities for building, deploying, and operationalizing ML systems in the context of ML engineering development for dealing with practical ML applications~\cite{kreuzberger2023machine,salama2021practitioners}. MLOps contains an automated and streamlined ML process, which not only benefits in-production ML model deployment but also facilitates managing risks when scaling wide-range ML applications to more practical use cases in various and complex real-world environments. 
Hence, MLOps covers data engineering, ML engineering with ML development and deployment, as well as application engineering with ML lifecycle management. 

Google has published a white paper in 2021 to provide an overview of MLOps~\cite{salama2021practitioners}, which identifies seven integrated and iterative components in the MLOps lifecycle: (a) data and model management; (b) ML development; (c) training operationalization; (d) continuous training; (e) model deployment; (f) prediction serving; and (g) continuous monitoring. They highlighted the central importance of the data and model management stage, which plays a vital role in governing ML artifacts from both data and model engineering perspectives. 
Besides, there are some MLOps tools that can be used to facilitate current research from the general framework (\eg, Kubeflow~\cite{kubeflow}, Amazon SageMaker~\cite{sagemaker}) to customized solutions for certain applications (\eg, Amazon Neptune~\cite{neptune}).
In summary, the overall objective of MLOps is productionizing machine
learning systems with the cooperation of data operations (DataOps) and development operations (DevOps).

\vspace{1mm}\noindent
\textbf{From MLOps to Graph MLOps?} As a prevalent and typical structural data, graph MLOps not only requires all the above-mentioned general components, but also demands unique characteristics specific to the graph MLOps lifecycle. 
First, from the perspective of data-centric graph ML in graph MLOps, graph data involves complex node and edge interactions with diverse characteristics, requiring corresponding graph-structured data storage methods and processing pipelines. The pipelines should be extensible to cater to different data sources and processing steps.
Besides, different from well-developed tabular, image, or text data, graph data quality is still an active research topic in graph MLOps, and enabling the data quality libraries to incorporate graph database representations remains an open question.
Second, from the perspective of the graph ML models, exploring the explainability of developed and deployed graph ML models is essential to model understanding and applications in real-world scenarios. 
Third, from the perspective of the feature store, graph MLOps require well-captured features to be stored and tracked, which contain comprehensive nodes, edges, and their relationships due to complex and diverse graph structural characteristics.
Finally, from the perspective of graph ML model inference, graph MLOps requires both neighbor information and node contexts to complete batched or streamlined inference in a model's online service.

\section{Future Directions}\label{sec:futures}

Data-centric graph machine learning (DC-GML) is an emerging and rapidly evolving research topic. Despite the significant progress that has been made, there are still numerous open problems that need to be addressed. In this section, we will explore several promising research directions in DC-GML.

\vspace{1mm}\noindent
\textbf{Exploration of complex and dynamic graph data.} 
Existing studies in DC-GML primarily concentrate on improving and exploiting simple graph data, characterized by attributes, staticity, homophily, and homogeneity. However, real-world graph data encompass a much broader range of properties, exhibiting diverse and complex properties (\eg, dynamic, heterophilic, and heterogeneous)~\cite{skarding2021foundations, wang2022survey, zheng2022graph}. One such example is dynamic graphs, where features and structures evolve over time, giving rise to their temporal nature. 
To effectively support data-centric machine learning on these complex and diverse graphs, it is essential to develop more targeted enhancement and exploitation strategies~\cite{guo2019attention}. 
Future studies should focus on exploring and devising approaches that are specifically tailored to handle the unique characteristics and challenges in terms of the complexity and diversity of graph data.
These strategies can involve capturing and modeling temporal dynamics~\cite{skarding2021foundations}, heterogeneity~\cite{pan2023unifying}, and heterophily~\cite{zheng2022graph}. 
By delving into these specific enhancement and exploitation strategies, researchers can advance the capabilities of DC-GML in handling complex and diverse real-world graph data. 

\vspace{1mm}\noindent
\textbf{General and automatic graph data improvement.} 
In Sec.~\ref{sec:tasks}, we have observed that most existing DC-GML works focus on addressing graph data availability or quality issues to enhance graph data in isolation. Specifically, they typically tackle one aspect of graph data, such as its structure, features, labels, or size~\cite{jin2020graph, jin2021heterogeneous, jin2020gcoarse, li2018deeper}. 
However, in real-world scenarios, graph data deficiencies often occur simultaneously across multiple aspects and are interconnected~\cite{huo2023t2, liu2023learning}. 
Consequently, current single-aspect methods are unable to adequately tackle the comprehensive array of deficiencies present in graph data.
Hence, there is a pressing demand for a comprehensive solution that can address data deficiencies across multiple aspects in a unified manner. This holistic approach should be capable of simultaneously enhancing graph structure, features, labels, and size, taking into account their interdependencies, as well as automatically detecting and identifying graph data deficiencies across various aspects. 
It would adaptively rectify these deficiencies, leading to enhanced data that is tailored to  specific shortcomings present in the dataset. By incorporating automatic detection and self-adaptive correction mechanisms, the solution can provide a more comprehensive and effective graph data improvement process. 

\vspace{1mm}\noindent
\textbf{Standardized graph data benchmarks.} 
Benchmarking is crucial for evaluating the effectiveness of DC-GML approaches. While there are numerous benchmark datasets available for model-centric graph learning, the focus on data-centric benchmarks has been comparatively limited~\cite{sen2008collective, shchur2018pitfalls, hu2020open}. Having robust and well-designed data-centric benchmarks is essential for a comprehensive evaluation of the data enhancement, exploitation, and maintenance processes in DC-GML. 
A high-quality data-centric benchmark provides researchers with a standardized platform to assess the performance of various approaches in enhancing and exploiting graph data. For instance, a benchmark dataset that incorporates natural structural deficiencies along with ground-truth completions can greatly aid in the development and evaluation of graph structure enhancement methods. By utilizing such a benchmark, researchers can test and compare different techniques in addressing structural deficiencies, thereby fostering innovation and advancement in the field. 
To sum up, the availability of diverse and representative data-centric benchmarks empowers researchers to gauge the effectiveness and generalizability of their proposed DC-GML methods. These benchmarks allow for objective comparisons and facilitate the identification of strengths and weaknesses in different approaches. Furthermore, they serve as a foundation for the establishment of best practices and standards within the DC-GML community.


\vspace{1mm}\noindent
\textbf{Collaborative development of graph data and model.}
In Sec.~\ref{sec:paradigm}, we have observed that  existing DC-GML learning strategies and methods fully exploit valuable information from graph data of limited-availability and low-quality. 
These methods resolve the graph data-centric challenges from a model perspective, performing as a bridge connecting current research on graph machine learning models towards the research centered on graph data engineering~\cite{liu2022graph, jin2021node, hu2020graph}.
A promising future direction in DC-GML is to further encourage more collaborative development and interaction between graph data engineering and graph machine learning models. 
For one thing, graph data improvement techniques help mitigate graph deficiencies, such as missing features or incomplete structure~\cite{spinelli2020missing};
Meanwhile, graph machine learning models can be refined to handle and leverage the enhanced graph data more effectively.
For another thing, graph machine learning models could facilitate graph data improvement, such as generative models for synthesizing augmented graph data~\cite{liu2023learning}.
For a typical example, GraphStorm~\cite{graphstorm}, a graph machine learning framework for enterprise use cases, simplifies the development, training, and deployment of graph learning models for industry-scale graph data. Meanwhile, it could deal with extremely large graphs (measured in billions of nodes and edges) with provided training and inference pipelines of graph models.
Hence, the co-development of graph data and models guarantees the optimization and customization of both aspects to various practical graph learning tasks. This ensures the enhancement of graph data availability/quality and the refinement of the graph model design/learning process, leading to improved performance in data-centric graph machine learning.

\vspace{1mm}\noindent
\textbf{Comprehensive graph data lifecycle management pipelines.} 
In this article, we have reviewed each phase of the graph data involved in the graph MLOps workflow modular by modular, encompassing graph data collection, exploration, improvement, exploitation, and maintenance. Real-world graph data often evolves rapidly, requiring continuous updates and management to ensure its availability and quality for specific graph learning tasks. 
This dynamic nature poses severe challenges to real-world DC-GML applications for seamlessly integrating all the above modular phases into a cohesive, dynamic, and holistic workflow. 
In light of these, a promising future direction of DC-GML is to develop dynamically-united graph data lifecycle management pipelines for effectively managing and leveraging graph data in a rapidly evolving environment.
It requires developing flexible and adaptable approaches that can accommodate the ever-changing demands and complexities of real-world graph data. 
An advanced graph data lifecycle management pipeline should stay up-to-date with the latest graph data and employ robust graph data processing practices. 
For instance, `Real-time Fraud Detection with Graph Neural Network'~\cite{RTFrau} is an end-to-end blueprint architecture for real-time fraud detection (leveraging graph database Amazon Neptune) using Amazon SageMaker~\cite{sagemaker} and Deep Graph Library (DGL)~\cite{wang2019dgl}.
It contains an end-to-end data processing pipeline to show how a real-world data pipeline could be, and it further constructs a heterogeneous graph from tabular data and trains a Graph Neural Network (GNN) model to detect real-time dynamic fraudulent transactions. 
Therefore, it can be anticipated that DC-GML can harness the power of real-world graph data under comprehensive graph data lifecycle management pipelines for enhanced decision-making and improved performance of graph machine learning models in various domains in the future.

\section{Conclusion}\label{sec:conclusion}
This article undertakes a comprehensive \textbf{review} and provides promising \textbf{outlook} for data-centric graph machine learning (DC-GML). 
We introduce a systematic framework with a thorough taxonomy of DC-GML that encompasses all stages of the graph data lifecycle, including graph data collection, exploration, improvement, exploitation, and maintenance. 
Through this article, we investigate three critical questions related to graph data availability and quality enhancement, learning, and the construction of graph MLOps systems from a graph-centric perspective. Additionally, we propose exciting future directions for the advancement of data-centric graph machine learning. It is anticipated that this work will foster a mutually beneficial collaboration and synergy between graph data engineering and graph model development within the realm of DC-GML.

\bibliographystyle{ACM-Reference-Format}
\bibliography{reference}



\end{document}